\pdfoutput=1

\documentclass{article}

\usepackage[main,final]{neurips_2026}

\usepackage[utf8]{inputenc} 
\usepackage[T1]{fontenc}    
\usepackage{hyperref}       
\usepackage{url}            
\usepackage{booktabs}       
\usepackage{amsfonts}       
\usepackage{nicefrac}       
\usepackage{microtype}      
\usepackage{xcolor}         

\usepackage{graphicx}
\usepackage{multirow}
\usepackage{enumitem}
\usepackage{amsmath}
\usepackage{cleveref}
\usepackage{makecell}
\usepackage{yuvion-neurips}

\hypersetup{
    colorlinks=true,
    linkcolor=red,
    urlcolor=cyan,
    citecolor=cyan,
}

\newcommand\textblue{\textcolor[RGB]{3, 111, 193}}
\newcommand\textorange{\textcolor[RGB]{255, 153, 0}}

\newcommand{\resultbest}[1]{\textorange{\textbf{#1}}}
\newcommand{\resultsecond}[1]{\textblue{\underline{#1}}}

\title{Why DiT Models Underperform as Representation Learners without Long Skip Connections}

\author{\parbox{14cm}{
    \centering
    \large{
        Benyuan Meng$^{1,2}$ \quad
        Yiliang Zhang$^{3}$ \quad
        Jin-Wen Wu$^{3}$ \quad
    }
    \\ 
    \large{
        Qianqian Xu$^{4,5}$\quad
        Qingming Huang$^{6,4,7}$\quad
        Longtao Huang$^{3}$\thanks{Corresponding author.}
    }
    \\
    \normalsize{
        \textnormal{$^1$Institute of Information Engineering, CAS}\\
        \textnormal{$^2$School of Cyber Security, University of Chinese Academy of Sciences}\\
        \textnormal{$^3$Alibaba Group}\\
        \textnormal{$^4$Key Lab. of Intelligent Information Processing, Institute of Computing Technology, CAS}\\
        \textnormal{$^5$Beijing Academy of Artificial Intelligence}\\
        \textnormal{$^6$School of Computer Science and Tech., University of Chinese Academy of Sciences}\\
        \textnormal{$^7$Key Laboratory of Big Data Mining and Knowledge Management, CAS}
    }
    \\
    \texttt{mengbenyuan@iie.ac.cn}\quad
    \texttt{\{suiyang.zyl,bage.wjw,kaiyang.hlt\}@alibaba-inc.com}\\
    \texttt{xuqianqian@ict.ac.cn}\quad  
    \texttt{qmhuang@ucas.ac.cn}
}}

\begin{document}

\maketitle
\pagestyle{yuvionheader}
\thispagestyle{yuvionheader}

\begin{abstract}
Diffusion feature — intermediate activations extracted from diffusion model backbones — has emerged as a promising approach for unifying generation and vision understanding (discrimination).
Despite the advancement of diffusion backbones into Diffusion Transformers (DiT), it has been found in previous literature that DiT models tend to yield weaker features despite their stronger generation capability.
This poses a threat to good generation-understanding unification.
In this work, we show via controlled from-scratch training that long skip connections (LSCs) affect feature quality: a DiT with LSCs can produce better features.
Furthermore, through detailed diagnosis, we reveal the mechanism causing this effect.
Specifically, the input to a diffusion backbone contains both the latent image and the diffusion noise, where the diffusion noise, though appearing in the form of noises, actually reflects the information of the diffusion trajectory.
To make the final prediction of a noisy variable ($v$ or $\epsilon$), the diffusion backbone needs both signals.
However, carrying both signals through the main backbone effectively ``pollutes'' the image signal with noises, leading to poor feature quality.
In comparison, LSCs can shortcut diffusion noises directly to where they are needed, leaving much weaker noise signals in the main backbone, thus improving feature quality.
To name the broader impact, this reveals that LSCs are not a mere trick but closely related to the diffusion objective.
The code is \href{https://github.com/Darkbblue/dit-lsc}{here}.
\end{abstract}

\section{Introduction}
\label{sec:intro}

Diffusion models are a trending type of generative models nowadays~\cite{ho2020denoising, dhariwal2021guideddiffusion, rombach2022high, podell2023sdxl}.
Beyond generation, it is also an important and emerging topic to explore the unification of generation and vision understanding tasks, such as semantic segmentation~\cite{DBLP:journals/corr/abs-2302-06378}, depth estimation~\cite{DBLP:conf/iclr/BochkovskiyDGSZ25}, or semantic correspondence~\cite{min2019spair} with diffusion models~\cite{baranchuk2021label, zhao2023unleashing, xu2023open, ke2023repurposing, DBLP:conf/cvpr/0001T0024a, li2023your, chen2024your}.
\textit{Diffusion feature} is an important sub-family of this research trend.
Every diffusion model \textit{pre-trained for generation}, \textit{e.g.}, Stable Diffusion v1.5~\cite{rombach2022high}, has a neural network backbone for $\epsilon$- or $v$-prediction~\cite{ho2020denoising, song2020score}, and many intermediate activations are created in the computation process.
If we extract these activations from the backbone, they could serve as versatile vision features, essentially similar to how people extract activations from ResNet~\cite{He16resnet} backbones as features.
High-quality diffusion features can enable better generation-understanding unification~\cite{wu2025qwenimagetechnicalreport, peng2026fgmllm}.

\begin{figure*}[t]
    \centering
    \includegraphics[width=\linewidth]{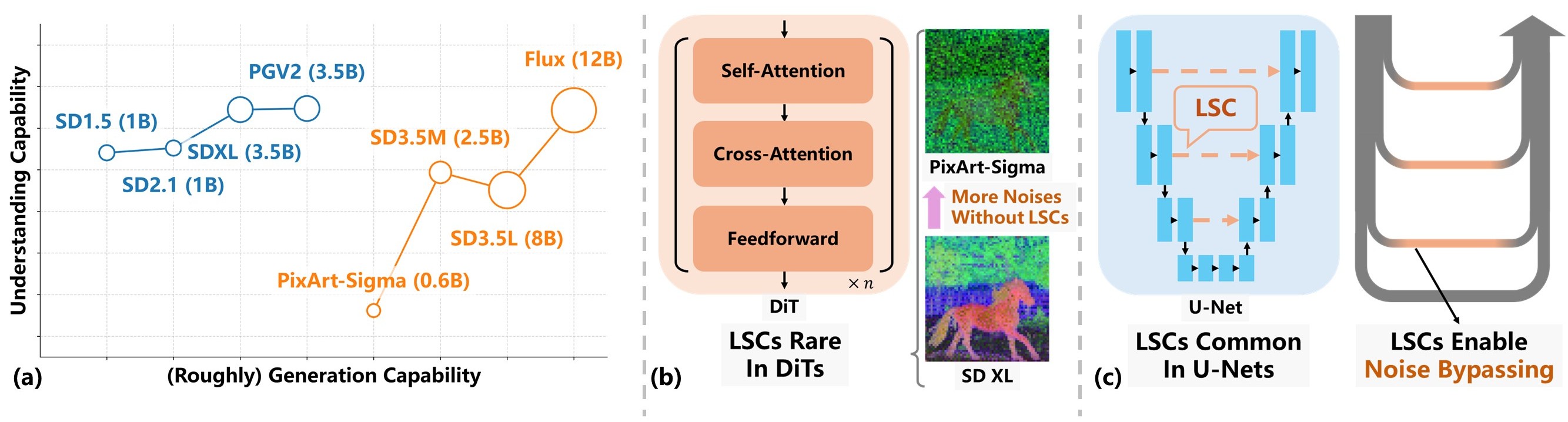}
    \caption{(a) Compared with U-Net diffusion models, DiT models yield diffusion features with relatively low understanding capability.
    (b) Most DiTs do not have LSCs, so the diffusion noise has to go through the target reconstruction section, causing features in this section to degrade.
    (c) We reveal a unique function of long skip connections (LSCs) in diffusion backbones: they work as shortcuts to transport the diffusion noise directly from the downsampling stage to the upsampling stage, thus \textit{bypassing} the middle section, where the most performative diffusion features can be extracted.}
    \label{fig:intro}
\end{figure*}

When studying diffusion features, the basic belief and hypothesis are that the advancement of generative capability naturally brings the advancement of understanding capability~\cite{my-SDXL}.
Here, the generative capability can be measured with FID score~\cite{fid}, CLIP score~\cite{DBLP:conf/emnlp/HesselHFBC21}, or user study~\cite{podell2023sdxl}, and the understanding capability can be measured using the performance on standard dense perception tasks~\cite{DBLP:journals/corr/abs-2302-06378, DBLP:conf/iclr/BochkovskiyDGSZ25, min2019spair}.
With this belief in mind, and considering that the mainstream architecture for diffusion models is turning to Diffusion Transformers (DiT)~\cite{Peebles23dit} from U-Nets, it is important that we can extract high-quality DiT features.
However, it is reported that DiT models tend to yield weaker features despite their stronger generation~\cite{ma} (\Cref{fig:intro}(a)).
This both poses a threat to generation-understanding unification and suggests a representation degradation phenomenon hidden behind the good generation, which may manifest one day and limit the ceiling of generation.

Notably, although the phenomenon is reported~\cite{ma}, the cause is still only preliminarily understood: the massive activation explanation appears to be applicable mostly to correspondence tasks (\Cref{sec:app-ma}), while the feature degradation exists universally (\Cref{sec:app-real-world}).
This indicates that the massive activation explanation is not exhaustive, and there should be a more universal cause.
On the other hand, many recent studies report that adding long skip connections (LSCs), a structure in U-Nets that shortcut information from the downsampling stage directly to the upsampling stage, can enhance DiT generation.
Inspired by this, we confirm via a controlled comparison experiment that the wide absence of LSCs from standard DiTs is a more universal cause, or at least part of it.

We do not stop at knowing LSCs are good but further ask \textit{why}.
It is common sense that such skip connections can carry high-frequency information like edges, and this is also used to explain the effectiveness of LSCs for DiT generation in some literature~\cite{u-vit}.
However, suppose LSCs only diverge some high-frequency semantics like edges from the middle part; why are features here improved simply by \textit{removing} some high-frequency semantics from them?
Through further hypothesizing and validating, we reveal a deeper mechanism that is fundamentally new compared to commonsense.
Specifically, \cite{my-SDXL} shows that diffusion backbones do not \textit{directly} predict a noisy variable from a noisy input, but instead reconstruct a semantic representation of the clean image $z_0$ in the middle of the network, based on which the final noise prediction is made.
To predict the final noisy variable, \textit{both} the reconstructed representation \textit{and} the diffusion noise carried in the noisy input should be transported to the final noise prediction section.
Here, the diffusion noise is \textit{not} a random noise: it appears in the form of noises but reflects certain information, \textit{i.e.}, the current denoising trajectory.
\textit{Without LSCs}, the only possible passage for the diffusion noise is through the entire target reconstruction section, thus ``\textit{polluting}'' the features extracted from this section, since it still takes the form of noises (\Cref{fig:intro}(b)).
\textit{With LSCs}, however, the diffusion noise can take the shortcut and thus bypass the target reconstruction section, maintaining high feature quality in this section (\Cref{fig:intro}(c)).
Hence, we add to the knowledge that LSCs can carry high-frequency semantics that they also carry diffusion noises.
This reveals that LSCs are not a mere trick to boost generation but are closely related to the diffusion objective.

In general, our contribution is as follows:
\begin{itemize}
    \item We provide one of the first systematic studies focusing on the discriminative quality of DiT features and notice a significant feature quality discrepancy between DiTs and U-Nets.
    \item We attribute the feature quality discrepancy to the LSC structure and provide a detailed, rigorously supported mechanistic explanation.
    \item Despite the initial representation learning motivation, we hope our mechanistic findings could make LSCs in DiTs a more widely adopted practice rather than an optional heuristic.
\end{itemize}

\section{Related Work}
\subsection{DiT Feature}
There are currently mainly three approaches to exploiting the understanding potential from a generative diffusion model: diffusion classifier~\cite{li2023your, chen2024your}, generative modeling~\cite{ke2023repurposing, DBLP:conf/cvpr/0001T0024a, wu2025qwenimagetechnicalreport}, and diffusion feature~\cite{baranchuk2021label, zhao2023unleashing, xu2023open, my-SDXL, tang2023emergent, ma}. Diffusion feature extracts intermediate activations from a diffusion backbone as features, regarding the diffusion backbone as an information bottleneck that compresses an image into a latent embedding.
These features are dense and rich in information, but not directly interpretable by humans.
Thus, a downstream model is required to process these features and produce the final predictions.

Most previous knowledge about diffusion feature is based on U-Net diffusion backbones.
\cite{ma} is the first to extend this methodology to DiT and surprisingly find that DiT features are generally worse.
The explanation proposed in \cite{ma} is \textit{massive activations}~\cite{ma-llm}, where the scale of a small number of elements in the feature is extremely large.

To some extent, diffusion feature can be viewed as a representation learning method.
However, the model is not directly trained for representation learning, but a generatively trained diffusion model is adapted to this task.
Compared to typical representation learning methods like DINO~\cite{dino}, diffusion feature performs especially well in scenarios where the background is relatively identical and the information is sparse~\cite{xie2025towards}.

\subsection{LSC-Augmented DiT Models}
Conventionally, long skip connections (LSCs) are a typical structure in U-Nets.
Since U-Nets adopt a downsampling-upsampling architecture, it is believed mandatory to utilize such skip connections to shortcut low-level/high-frequency semantics like edges, which will otherwise be lost during downsampling.
At first, when the diffusion model community shifts from U-Net backbones to DiT backbones, LSCs were naturally discarded, since DiTs have identical resolution across blocks.
However, quite a few recent studies find that adding LSCs to DiT can enhance generation~\cite{u-vit, skip-dit, u-dit}.
Compared to them, our findings are fundamentally different in two ways:
(i) We find LSCs are beneficial not only to generation but also to representation learning.
(ii) We reveal the \textit{diffusion noise} bypassing behavior of LSCs, which is different from the high-frequency semantic shortcutting~\cite{u-vit} and loss smoothing~\cite{skip-dit} explanations.

Additionally, there are also LSC-augmented DiT models with actual resolution changes~\cite{tian2024udits, tian2025urepa}.
To isolate the contribution of LSCs, we use \textit{isotropic} DiTs as the backbone to add LSCs onto.
For further comparison with models with actual resolution changes, please refer to \Cref{sec:app-unet-dit}.

\section{Preliminaries}
\label{sec:pre}
\subsection{Diffusion Models}
Based on \cite{flow-matching}, the diffusion process starts at a point sampled from the source distribution, progressively moving it towards the target distribution, where the two distributions are typically noises and images.
Each step during this process is called a timestep $t \in [0, 1]$.
At each timestep, a backbone network is used to predict the velocity of the movement.
We can denote this network as $v_\theta$, parameterized by $\theta$.
This network used to be U-Nets, but now it is mostly ViTs, known as DiTs.

To actually train the model, we need to sample an image $x_0$, its corresponding user prompt $c$, and a timestep $t$, calculating a ground truth based on them.
First, the image is compressed into the latent space by a pre-trained VAE as $z_0 = \text{VAE}(x_0)$, and we sample a Gaussian noise $z_1$.
Then, we have
\begin{equation}
\label{equ:v}
\begin{aligned}
    &z_t = tz_0 + (1-t)z_1\\
    &v_t = \frac{dz_t}{dt}=z_0-z_1
\end{aligned}    
\end{equation}
Afterward, we can train the backbone network to predict the velocity:
\begin{equation}
    \mathcal{L}=\mathbb{E}_{(z_0,c)\sim\mathcal{D},z_1,t}||v_\theta(z_t,t,c)-v_t||^2
\end{equation}
The formulation above is denoted as \textit{flow matching}, and the more conventional DDPM~\cite{ho2020denoising} and score matching~\cite{song2020score} frameworks can be seen as special forms of flow matching.
Notably, although the difference in the theoretical framework may not be that important, the prediction objective of the backbone model, \textit{i.e.}, $\epsilon$-, $v$-, or $x$-prediction, does matter.
In the main body of this manuscript, we will use $\epsilon$-prediction for experiments, and we will show in \Cref{sec:app-evx} how our findings extend to $v$- and $x$-prediction.

\subsection{Diffusion Feature}
To extract a feature from a diffusion backbone, we need an image, a selected timestep, and, optionally, a prompt describing the image content.
Denoting them as $x_0, t, c$, we have:
\begin{equation}
\begin{aligned}
    z_0 &= \text{VAE}(x_0) \\
    z_t &\sim q(z_t|z_0,t) \\
    v &= v_\theta(z_t, t, c)
\end{aligned}
\end{equation}
Respectively, the three lines indicate encoding the image into latent space with a VAE, sampling a noisy latent corresponding to the timestep $t$, and sending all three inputs into the diffusion backbone.
We discard the output $v$.
Instead, we gather some intermediate activations as features.
Such activations can be extracted from a variety of spots, such as the residual flow between two adjacent layers or the query vector used to compute the attention matrix in an attention layer~\cite{my-SDXL}.
In DiT models, these activations are patchified 1D features, and we can reshape them back to 2D features and resize them to the original image size using bilinear interpolation.

Obviously, there are many legal feature extraction spots, but they cannot all be harvested due to limited computational resources, so there is a need to select a few near-optimal activations as features.
To this end, a three-step strategy can be taken~\cite{my-SDXL}.
\begin{enumerate}[label=(\roman*)]
    \item Based on both PCA analysis and quantitative comparison within small ranges, study the properties of features in this backbone.
    \item Based on the understanding of feature properties, rule out low-quality features qualitatively.
    \item Perform a filtered quantitative comparison among the remaining features on a small and fast downstream task to find the most performative features. In this study, we adopt training-free semantic correspondence~\cite{min2019spair} as the task.
\end{enumerate}

After selecting activations as features, diffusion features can be used in a very straightforward way.
Basically, suppose the original input to a visual perception task is a raw image; we could replace the image with the corresponding diffusion feature.
Commonly, such replacement requires the dedicated training of a compatible downstream task head~\cite{baranchuk2021label}, and we could fine-tune the diffusion backbone on the task for better performance~\cite{zhao2023unleashing}, or, in much rarer cases, use diffusion features in a totally training-free manner~\cite{zhang2023tale}.

\section{Long Skip Connections Matter}
\label{sec:lsc}

\subsection{Core Hypothesis}

\begin{figure}[t]
    \centering
    \includegraphics[width=.8\linewidth]{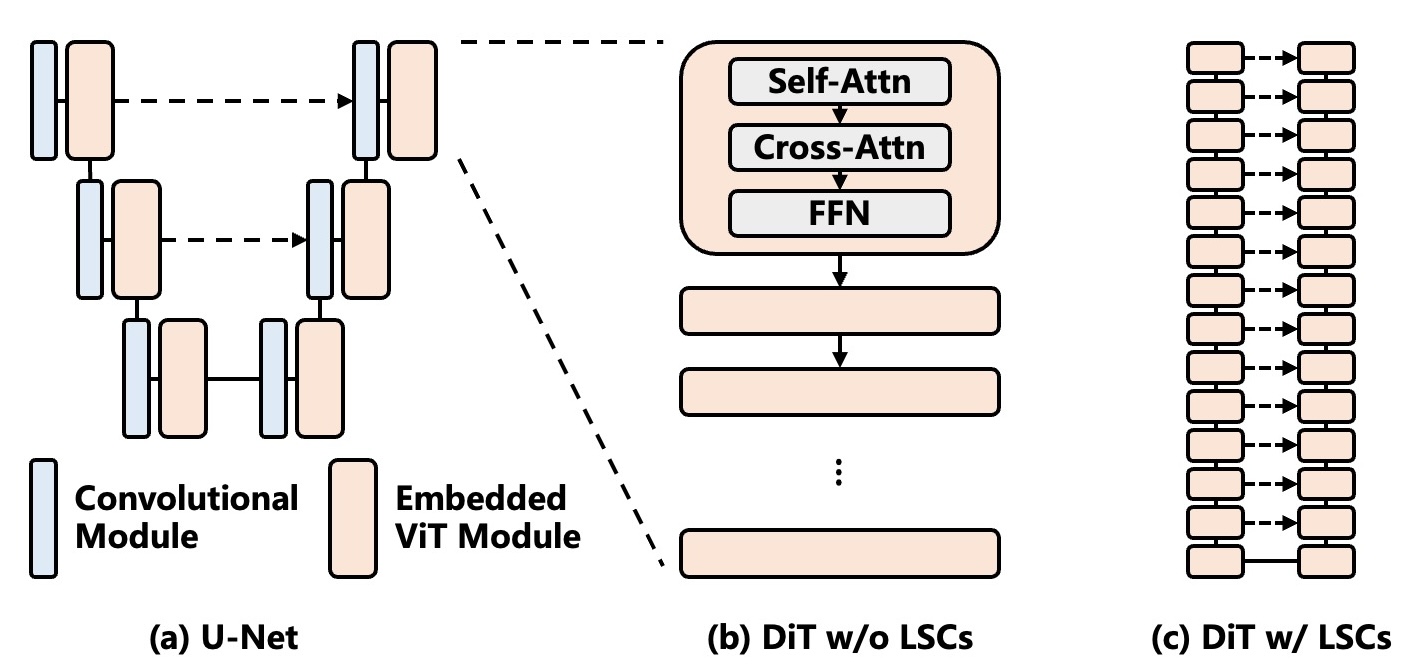}
    \caption{(a) A typical U-Net architecture of modern diffusion models is not purely convolutional but contains many large embedded ViT modules.
    (b) A DiT backbone can, from a certain perspective, be viewed as extracting an embedded ViT module and greatly extending it.
    (c) We can fold the DiT chain in half to form an LSC-augmented DiT.}
    \label{fig:lsc-dit-structure}
\end{figure}

With the recent emergence of LSC-augmented DiTs~\cite{u-vit, skip-dit, u-dit}, it gradually becomes acknowledged that LSCs are beneficial for DiT \textit{generation}.
Furthermore, the basic belief of diffusion feature studies tells us a stronger generation model should generally lead to stronger features.
Hence, it is natural to hypothesize that LSCs are beneficial for DiT \textit{features} as well.
For a visual illustration of the architectures of U-Net and DiT diffusion models, refer to \Cref{fig:lsc-dit-structure}(a,b).

As preliminary evidence, we test on pre-trained DiT-XL/2~\cite{Peebles23dit} and U-ViT~\cite{u-vit} models where the generative capability roughly aligns.
The result that U-ViT (with LSCs) yields significantly better features than DiT-XL/2 (without LSCs) (53.53 v.s. 44.01 on SPair (a)) implies the benefit of LSCs.
Nevertheless, the full results in \Cref{sec:app-pretrained} also suggest a controlled experiment to exclude the confounders.

\subsection{Hypothesis Validation}
We further conduct a controlled from-scratch training experiment to validate our hypothesis above.
Basically, we train two DiT~\cite{Peebles23dit} models from scratch, with the only difference being whether there are LSCs or not, and compare them on six downstream benchmarking settings.

\noindent \textbf{Training Dataset.}
We train our models on ImageNet~\cite{imagenet} 256$\times$256.
This dataset contains 1,281,166 images in 1000 classes, all in 256$\times$256 resolution.

\noindent \textbf{Models.}
We choose the DiT-XL/2 option in \cite{Peebles23dit}.
The only modification we make to the models is adding LSC structures.
The LSC structure we use is formulated as
\begin{equation}
\label{equ:lsc}
    x \leftarrow \text{Linear}(\text{Norm}(\text{Concat}(x, \text{skip})))
\end{equation}
where $x$ is the inter-layer residual flow activation, $\text{skip}$ is the activation shortcut here through LSCs, Linear is a single fully connected layer, Norm is LayerNorm, and Concat is concatenation along the channel dimension.
A standard DiT architecture is linear, and intuitively, our modification folds the architecture in half from the middle to enforce it into a conceptual ``U'' shape.
For a visual illustration, please refer to \Cref{fig:lsc-dit-structure}(c).

\noindent \textbf{Training Details.}
We simply keep all hyperparameters in the original script unchanged, using a global batch size of 256.
Additionally, both models are trained for 400k steps by default.
To further isolate the contribution of LSCs from the models' generative capability, we also provide a \textit{w/ LSC 100k} setting trained for fewer steps to align the FID scores.

\noindent \textbf{Evaluation.}
For the generation capability, we use FID scores as the metric, measured under CFG=1.5.
We use six benchmarking settings for the discriminative performance, which are listed in \Cref{tab:setting}.
\begin{enumerate}[label=(\roman*)]
    \item \textit{SPair (a)} to \textit{SPair (c)}: these use SPair-71k~\cite{min2019spair} dataset as the benchmark. The difference between the three settings is the parameters trainable on the downstream task: \textit{SPair (a)} is training-free, \textit{SPair (b)} trains only a downstream head, and \textit{SPair (c)} fine-tunes the DiT model.
    \item \textit{Horse-21} and \textit{CityScapes}: these use semantic segmentation task, one on Horse-21~\cite{baranchuk2021label} dataset and the other on CityScapes~\cite{Cordts16city} dataset, both only training a downstream head.
    \item \textit{Depth}: this uses depth estimation task on NYU Depth v2~\cite{nyudepthv2}, and we fine-tune the DiT model on this setting.
\end{enumerate}
Unlike in generation, classifier-free guidance (CFG) is not commonly used for feature extraction.

\begin{table}[t]
\centering
\caption{Comparison of the six benchmarking settings we use throughout this study.}
\label{tab:setting}
\begin{tabular}{@{}lcccc@{}}
\toprule
Setting Name & Task                    & Dataset      & Metric  & Trainable Params \\ \midrule
SPair (a)    & Semantic Correspondence & SPair-71k    & PCK@0.1 & None             \\
SPair (b)    & Semantic Correspondence & SPair-71k    & PCK@0.1 & Downstream Head  \\
SPair (c)    & Semantic Correspondence & SPair-71k    & PCK@0.1 & All              \\ \midrule
Horse-21     & Semantic Segmentation   & Horse-21     & mIoU    & Downstream Head  \\
CityScapes   & Semantic Segmentation   & CityScapes   & mIoU    & Downstream Head  \\ \midrule
Depth        & Depth Estimation        & NYU Depth v2 & d1      & All              \\ \bottomrule
\end{tabular}
\end{table}

The results are shown in \Cref{tab:lsc-exp}.
We can observe that \textit{w/ LSC} is significantly stronger than \textit{w/o LSC} across all settings, even for \textit{w/ LSC 100k} where the FID scores are aligned.
Thus, we could causally confirm that \textit{LSCs are indeed an important contributing factor}.
More importantly, even \textit{w/ LSC 100k} outperforms \textit{w/o LSC}, where the two models have similar generation capability.

\begin{table*}[t]
\centering
\caption{Discriminative performance evaluated on all six settings as listed in \Cref{tab:setting}. Plus, the FID scores are obtained under CFG=1.5. Best/second marked as \textorange{\textbf{bold}} and \textblue{\underline{underlined}}.}
\label{tab:lsc-exp}
\begin{tabular}{@{}lccccccc@{}}
\toprule
LSC Setting & FID      & SPair (a) & SPair (b) & SPair (c) & Horse-21 & CityScapes & Depth \\ \midrule
w/o LSC & 20.87         & 34.68      & 52.62      & \resultsecond{64.70}      & 30.35      & 50.22      & 0.70       \\
w/ LSC  & 14.39   & \textorange{\textbf{41.10}}      & \textorange{\textbf{60.68}}      & \textorange{\textbf{65.69}}      & \textorange{\textbf{36.32}}      & \textorange{\textbf{51.55}}      & \textorange{\textbf{0.74}}       \\ 
w/ LSC 100k & 21.67         & \resultsecond{38.85}      & \resultsecond{57.36}      & 64.43      & \resultsecond{35.24}      & \resultsecond{50.70}      & \resultsecond{0.73}      \\ \bottomrule
\end{tabular}
\end{table*}

\begin{figure}[t]
    \centering
    \includegraphics[width=.85\linewidth]{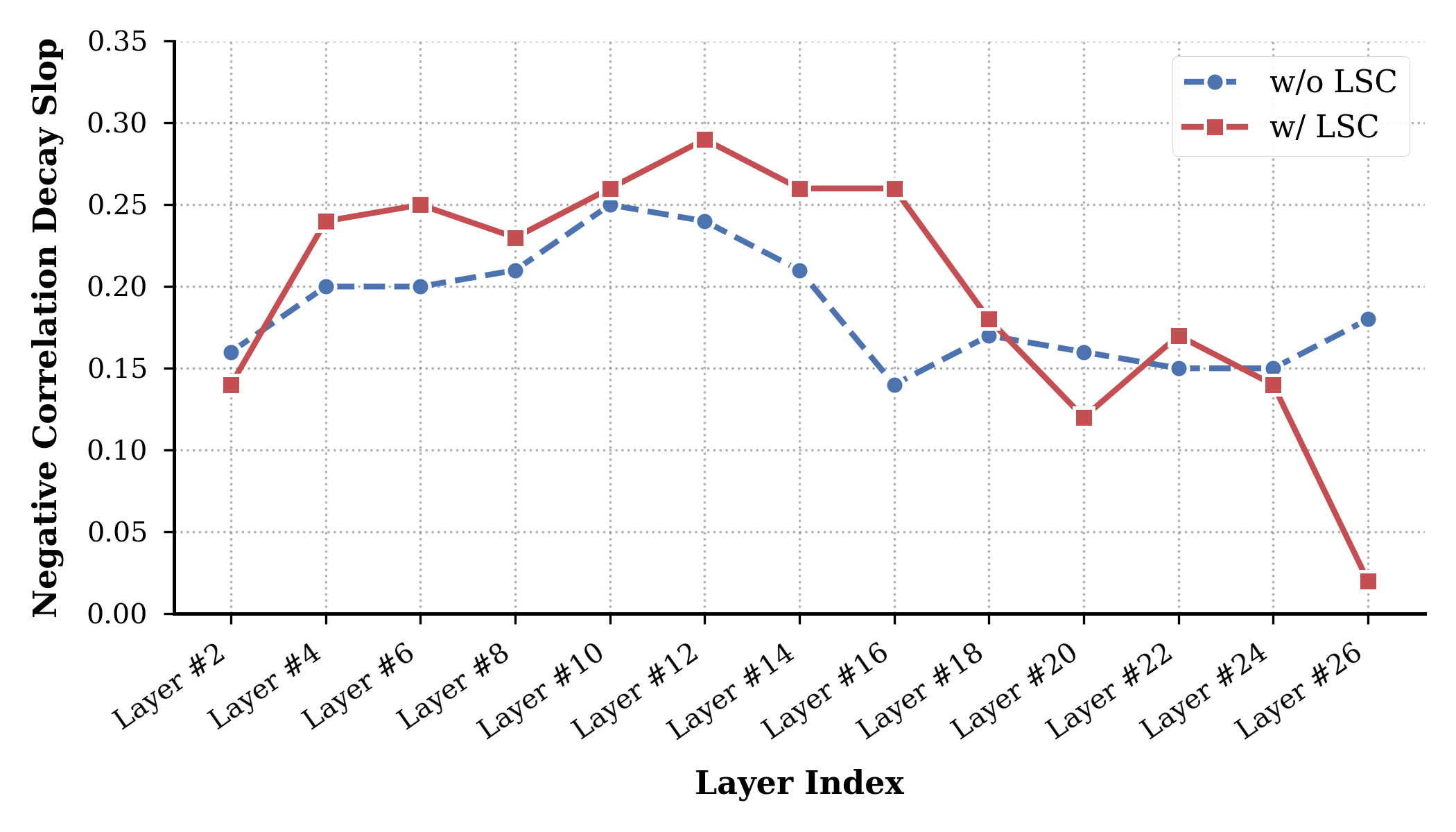}
    \caption{Correlation Decay Slope measuring results, where we report the negative slope of the linearly regressed lines.}
    \label{fig:cds}
\end{figure}

For additional information, we also adopt Correlation Decay Slope (CDS)~\cite{cds} for evaluation.
The results in \Cref{fig:cds} are the negative slopes, where a moderate slope indicates stronger feature expressiveness, while a slope that is either too low or too high is not favored.
We can observe that in middle layers (12-16), \textit{w/ LSC} shows higher slope values compared to \textit{w/o LSC}.
This supports the benefit of LSCs.

\section{WHY Long Skip Connections Matter}

\subsection{Why Do We Not Fully Understand LSCs Yet}
At first glance, the mechanics behind LSCs' contribution should be already known.
However, we will show that two competing explanations cannot fully explain our previous experiments.

(i) Feature enhancement does not solely stem from better generation.
In \Cref{tab:lsc-exp}, even when we manually align the generation capability with \textit{w/ LSC 100k}, the feature quality is still better than \textit{w/o LSC}.
(ii) LSCs shortcutting high-frequency semantics (\textit{i.e.}, edges)~\cite{u-vit} cannot fully explain the observation.
Since LSCs can shortcut high-frequency semantics, the features extracted from the middle part should have more concentration of low-order semantics, which means they might perform better for coarse-grained tasks like image classification.
However, what we have tested the models on are fine-grained tasks such as semantic segmentation.
Hence, there should be some additional mechanisms that co-govern the gain brought by LSCs.

\subsection{Target Reconstruction Behavior}
Our hypothesis to explain why LSCs have such a positive effect is based on a finding in \cite{my-SDXL}.
Here, we first re-formulate the ``diffusion noise'' property found in \cite{my-SDXL} as ``target reconstruction''.
Note that we do \textit{not} claim originality for this subsection's content.
It is here to smooth the narrative later.

Based on the flow matching formulation (\Cref{sec:pre}), both the input variable and the predicted variable of the diffusion backbone network are noisy variables.
A natural interpretation of this would be that diffusion activations should be mostly noisy.
However, the diffusion noise property states that, in the middle section of the network, intermediate activations are rather clean from noises.
Actually, these activations are an important source of high-quality diffusion features.
Now, considering that diffusion features extracted from the middle section in fact reflect the semantics of the target image $z_0$, this middle section's role is to reconstruct an internal representation for the target image based on the noisy input $z_t$.
Then, the network could encode its final prediction based on this reconstructed target representation.
In conclusion, this target reconstruction behavior shows that diffusion backbones take \textit{two rather than one} step to make their final prediction.

To facilitate the narrative, we conceptually divide the diffusion backbone into three coarse sections based on their roles, which are indicated by three different colors, red, green, and blue, in \Cref{fig:three-stage}:
\begin{enumerate}[label=(\roman*)]
    \item \textbf{Noise suppression}: To prepare for the later acquisition of the target representation, the noises in the inputs are gradually refined and suppressed to some extent.
    \item \textbf{Target reconstruction}: Next, the backbone reconstructs a latent representation of the target image latent within activations.
    \item \textbf{Noise prediction}: Finally, the model predicts the velocity, which is a noisy quantity.
\end{enumerate}

\subsection{Diffusion Noise Bypassing Hypothesis}

Next, we develop upon this target reconstruction behavior to form our diffusion noise bypassing hypothesis.
Consider the flow matching formulation: $v_t = z_0-z_1$.
We hypothesize that, to make the final prediction, \textit{both} the target information ($z_0$) \textit{and} the diffusion noise ($z_1$) are required, just like how both the starting and terminating points are required to determine a direction.
Of the two, the target information is provided by the target reconstruction section, and the diffusion noise is carried within the input variable.

\begin{figure*}[t]
    \centering
    \includegraphics[width=.9\linewidth]{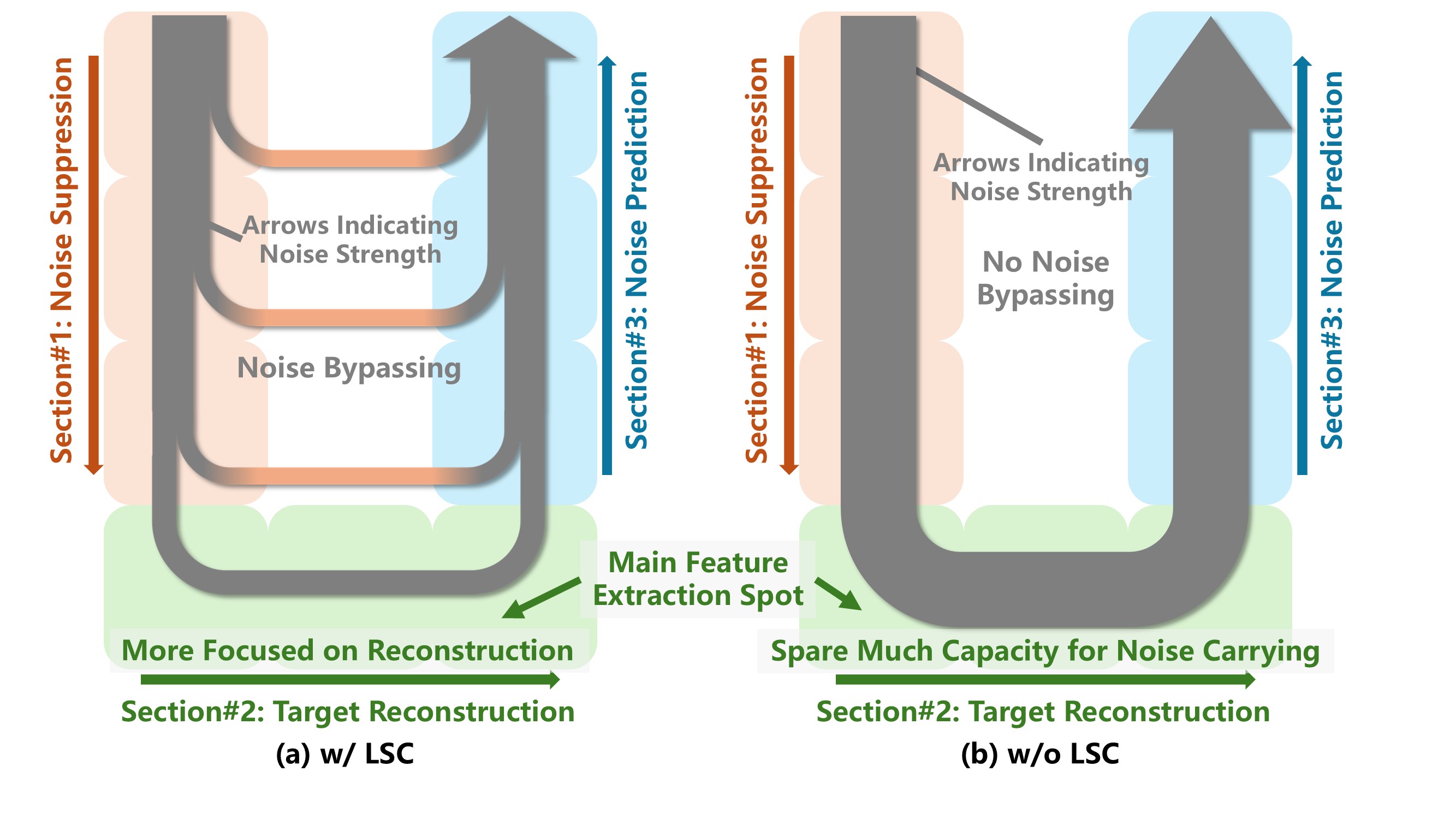}
    \caption{The noise bypassing hypothesis, where we use gray arrows to indicate noises and their \textit{width as noise strength}. With LSCs, the diffusion noise is shortcut through LSCs, leaving only weak noises into the target reconstruction section. Without LSCs, the diffusion noise is carried through the target reconstruction section, leading to feature degradation.}
    \label{fig:three-stage}
\end{figure*}

Furthermore, in order for the noise prediction section to make the prediction, both target and diffusion noise signals should be accessible to the noise prediction section.
Since the target reconstruction section and the noise prediction section are directly adjacent, the target information can be transported to the noise prediction section simply through the main residual flow.
However, how the diffusion noise is routed is where LSCs play their role.
\textit{Without LSCs}, as shown in \Cref{fig:three-stage}(right), the diffusion noise must also go through the main residual, but there is a target reconstruction section in between the noise suppression and noise prediction sections.
Therefore, the activations in the target reconstruction section have to spare some capacity to carry the diffusion noise, which makes their original goal, \textit{i.e.}, target reconstruction, degrade.
\textit{With LSCs}, as shown in \Cref{fig:three-stage}(left), however, the diffusion noise can directly take the shortcut and \textit{bypass} the target reconstruction section, thus maintaining strong feature quality in the target reconstruction section.
Thus, we make a hypothesis that could explain why LSCs can improve diffusion feature quality.

\subsection{Hypothesis Validation}

\begin{figure}[t]
    \centering
    \includegraphics[width=1\linewidth]{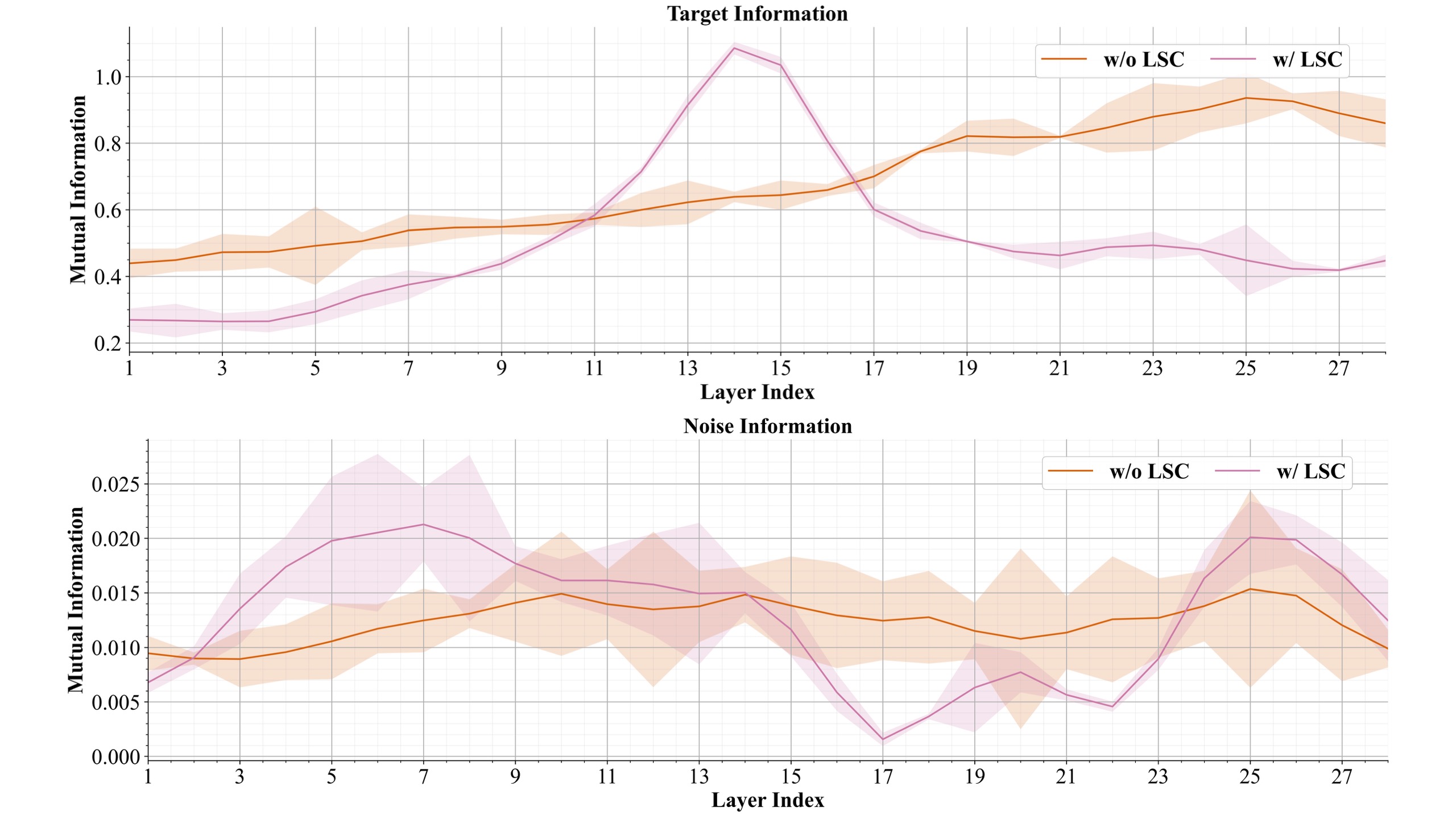}
    \caption{Mutual information between each layer's activation and target/diffusion noise, measured using MINE. Shadows indicate variance across three seeds.}
    \label{fig:mech-mi}
\end{figure}

First, we use mutual information probing to actually measure the amount of the target information and diffusion noise described in the noise bypassing hypothesis, hoping to provide \textit{correlation} evidence.
The target information contained in each activation is defined as the mutual information between this activation and the target latent ($z_0$), and the diffusion noise contained in each activation is defined as the mutual information between this activation and the noise added to the latent image ($z_1$).
To measure the mutual information, we adopt the Mutual Information Neural Estimator (MINE)~\cite{mine}, with a setting of (MINE hidden channel as 512, projection head channel as 256, training steps as 600).
The results are plotted in \Cref{fig:mech-mi}, and we can see that LSCs can indeed reduce the amount of diffusion noise and increase the amount of target information in the target reconstruction section (roughly the middle section).
We also provide an alternative probing using a different MINE setting in \Cref{sec:app-mi}.

\begin{table}[t]
\centering
\caption{Causal intervention experiment to separate the effect of high-frequency semantics and diffusion noises. All settings are trained for 400k steps before evaluation.}
\label{tab:causal}
\begin{tabular}{@{}lccccccc@{}}
\toprule
Setting              & Loss@1k & Loss@5k & FID    & SPair (a) & SPair (b) & SPair (c) & Horse-21 \\ \midrule
w/o LSC              & -       & -       & 20.87  & 34.68     & 52.62     & 64.70     & 30.35    \\
w/ LSC               & 0.1744  & -       & 14.39  & 41.10     & 60.68     & 65.69     & 36.32    \\ \midrule
$z_0^{\text{high}}+\epsilon$   & 0.1189  & 0.0705  & NaN    & NaN       & NaN       & NaN       & NaN      \\
$z_0^{\text{high}}+\epsilon'$ & 0.1309  & 0.1222  & 37.04  & 26.76     & 44.43     & 58.43     & 22.59    \\
${{z_0}'}^{\text{high}}+\epsilon$ & 0.1196  & 0.1023  & 306.92 & 20.68     & 39.65     & 50.97     & 18.67    \\ \bottomrule
\end{tabular}
\end{table}

Next, to provide relatively \textit{causal} evidence, we conduct a controlled study.
Here, what we already know is that LSCs mainly carry high-frequency information.
What we want to validate is that LSCs carry not only high-frequency semantics but also diffusion noises.
The experiments we design include three settings besides the baseline:
\begin{enumerate}[label=(\roman*)]
    \item Replacing the $skip$ variable in \Cref{equ:lsc} with the linear addition of the high-frequency part of the original latent ($z_0^{high}$) and the original noise ($z_1$). This serves as a replacement baseline.
    \item Compared to setting (i), replacing the original noise with a random noise ($z_1'$). This isolates the contribution of high-frequency semantics.
    \item Compared to setting (i), replacing the original latent with a latent from a different image (${z'}_0^{high}$). This isolates the contribution of diffusion noises.
\end{enumerate}
We can observe from \Cref{tab:causal} that:
\begin{enumerate}[label=(\roman*)]
    \item All three settings show smaller losses yet weaker representation \textit{and} generation performance compared to \textit{w/ LSC}.
    Setting (i) even collapses to NaN, and Setting (ii) fails to converge (very high FID scores).
    This is a sign of representation collapse: injecting external signals creates some information shortcut and thus trivializes the task, making training worse.
    \item Settings (ii) and (iii) both show different degradation severity compared to setting (i), meaning that removing either of the two signals will affect the training process.
    This further confirms that both high-frequency semantics and diffusion noises play a non-negligible role in making LSCs work.
\end{enumerate}
Nevertheless, since the training process is highly complex, we cannot say for sure from the results that diffusion noises are more important than high-frequency semantics or vice versa.
In conclusion, LSCs actually carry a \textit{mixture} of high-frequency semantics and diffusion noises, where both of them are non-negligible.
This effectively extends our previous knowledge that only covers high-frequency semantics but not diffusion noises.

\subsection{Discussion}
Our diffusion noise bypassing hypothesis serves as knowledge of the inner workings of diffusion model backbones.
Based on this new knowledge, we can see that LSCs can enable diffusion backbones to work in a more optimized way: the activations in the target reconstruction section no longer need to spare much capacity to carry the diffusion noise.
Hence, \textit{even if one's goal is not related to features, but only generally better diffusion models, we believe LSCs are still beneficial}.
For example, Skip-DiT~\cite{skip-dit} and U-ViT~\cite{u-vit} extensively show that LSCs can enhance \textit{generation}, where we believe there is a certain causal relation.
To be specific, REPA~\cite{REPA} finds that the quality of inner representations in diffusion backbones is essential to generation, where better representations lead to stronger generation and faster convergence.
Hence, it is likely that the stronger generation brought by LSCs \textit{partially} stems from the stronger feature they enable.
Furthermore, compared to explicitly aligning diffusion features to an external representation model in REPA, LSCs introduce less constraint and can thus better exploit the advantage of the diffusion training objective itself.

\section{Conclusion \& Limitation}
\label{sec:conclusion}
This work reveals that long skip connections (LSCs) are beneficial for DiT, with regard to not only generation but also feature quality.
This can also explain why current large-scale DiT models tend to yield weaker features than U-Net diffusion models, despite their stronger generation capability.
More importantly, we propose a mechanistic hypothesis termed \textit{diffusion noise bypassing} to explain why LSCs have such an effect.
In general, we show that LSCs are not merely an engineering trick, but actually have a close relation to the diffusion process itself.
Thus, we hope this work can persuade more researchers and developers to consider adding LSCs back to the backbone models in the DiT era.

Unfortunately, we were not able to effectively augment existing pre-trained DiT models with LSCs, which has been found to be very difficult (\Cref{sec:add-lsc}).
Additionally, our controlled experiments are based on the conventional DiT architecture rather than the MMDiT architecture, which remains underexplored.
We also do not claim exhaustiveness of our LSC explanation as the cause of DiT feature degradation.

\section*{Acknowledgments}
This work was supported in part by National Natural Science Foundation of China: 62525212, U23B2051, 62236008, 62441232, 62521007 and U21B2038,
in part by Youth Innovation Promotion Association CAS,
in part by the Strategic Priority Research Program of the Chinese Academy of Sciences,
Grant No. XDB0680201,
and in part by the project ZR2025ZD01 supported by Shandong Provincial Natural Science Foundation.

\bibliography{neurips_2026}

\newpage
\appendix

\section{Appendix}

\subsection{Relation to Massive Activation Suppression}
\label{sec:app-ma}
In the original \cite{ma}, massive activations are found to be harmful to DiT feature quality.
However, they mostly focus on correspondence tasks, which conventionally use training-free nearest neighbor for prediction, leading to a natural sensitivity to massive activations.
Here, we evaluate the massive activation suppression method proposed in \cite{ma} on the semantic segmentation task, with the results in \Cref{tab:app-ma}.
It can be observed that the performance gain is much smaller on semantic segmentation than on semantic correspondence.
This suggests that massive activations are not that harmful to tasks beyond semantic correspondence, especially when a trained downstream head that can refine the features is involved.
Additionally, comparing the results on SPair (a), we can see that when adding massive activation suppression to \textit{w/o LSC} and \textit{w/ LSC}, their performance is both improved, with the gap between them remaining.
This indicates that the effects of massive activation suppression and LSCs are roughly orthogonal.

\begin{table}[t]
\centering
\caption{Comparing the massive activation suppression method (MAS) proposed in \cite{ma} on different tasks.}
\label{tab:app-ma}
\begin{tabular}{@{}lcc@{}}
\toprule
LSC Setting   & SPair (a)                 & Horse-21                  \\ \midrule
w/o LSC       & 34.68                     & 30.35 \\
w/o LSC + MAS & 36.32 & 31.27                     \\
w/ LSC        & 41.10                     & 36.32                     \\
w/ LSC + MAS  & 44.55 & 36.51                     \\ \bottomrule
\end{tabular}
\end{table}

As additional evidence, we also measure the massive activation phenomenon when LSCs are added.
From \Cref{tab:app-ma2}, we can observe that LSCs have a slight effect on suppressing massive activations, but the effect is limited.
This shows LSCs have additional, separate mechanics.

\begin{table}[t]
\caption{Measuring the massive activation phenomenon without and with LSCs.}
\label{tab:app-ma2}
\begin{tabular}{@{}lccc@{}}
\toprule
Setting & Channel Scale Median & Channel Scale Mean & Channel Scale Top5                    \\ \midrule
w/o LSC & 7.36                 & 8.41               & 555.50, 386.00, 302.00, 282.75, 50.43 \\
w/ LSC  & 7.32                 & 10.82              & 273.75, 233.12, 217.00, 141.12, 91.56 \\ \bottomrule
\end{tabular}
\end{table}

In general, massive activation as the cause of DiT feature degradation is not exhaustive.

\subsection{DiT Models with Actual Resolution Changes}
\label{sec:app-unet-dit}
We wish to first clarify the two factors: LSCs and resolution changes.
LSCs can be applied to \textit{both} isotropic DiTs \textit{and} DiTs with resolution changes.
However, a DiT backbone with resolution changes has to utilize LSCs to preserve information that would otherwise be lost during downsampling.
Hence, the two factors are only partially entangled, thanks to the transition and wide adoption of the transformer architecture~\cite{tian2023spi, zhang2024dnd}.

For comparison, we modify the DiT~\cite{Peebles23dit} code to implement resolution change in a minimal way, leading to two alignment options:
(i) Down-Up-DiT-Param: this setting roughly aligns the parameter size to the DiT-XL/2 option we use in the main manuscript.
(ii) Down-Up-DiT-Flops: this setting has more parameters, but the overall FLOPs roughly align to the DiT-XL/2 option.
Besides the two alignment options, we also provide two downsampler/upsampler design variations:
(i) Convolutional-based;
(ii) Swin-Transformer-style~\cite{swin-transformer} patch merging/expanding design.
This leads to four variations in total, and we compare all of them against \textit{w/ LSC} on generation and all six benchmark settings in \Cref{tab:setting}.

The results are shown in \Cref{tab:app-du}.
We can observe roughly two effects from resolution changes:
(i) By forcing the backbone to actively utilize LSCs via resolution downsampling, the feature quality tends to improve.
(ii) However, as the middle-layer features have a lower resolution, they are not that suitable for fine-grained tasks such as semantic segmentation, leading to a quality drop.
Overall, there are two effects that fall in opposite directions, and the general effect relies on what task the features are used for.

\begin{table}[t]
\centering
\caption{Generation and discrimination performance of DiT models with actual resolution changes. Best and second marked as \textorange{\textbf{bold}} and \textblue{\underline{underlined}}.}
\label{tab:app-du}
\begin{tabular}{@{}lccccccc@{}}
\toprule
Model      & FID            & SPair (a)      & SPair (b)      & SPair (c)      & Horse-21       & CityScapes     & Depth         \\ \midrule
w/ LSC     & \resultbest{14.39} & 41.10           & \resultsecond{60.68}    & 65.69          & \resultbest{36.32} & \resultbest{51.55} & 0.74          \\ \midrule
Param-Conv & 21.31          & 38.62          & 55.83          & 65.04          & 31.62          & \resultsecond{45.90}     & \resultsecond{0.76}    \\
Flops-Conv & 19.09          & 39.04          & 58.31          & 67.03          & \resultsecond{32.05}    & 45.84          & \resultbest{0.79} \\
Param-Swin & 18.53          & \resultsecond{44.35}    & 60.65          & \resultsecond{68.22}    & 28.02          & 43.98          & \resultsecond{0.76}    \\
Flops-Swin & \resultsecond{16.20}     & \resultbest{45.96} & \resultbest{62.52} & \resultbest{69.56} & 29.81          & 45.27          & 0.75          \\ \bottomrule
\end{tabular}
\end{table}

\subsection{Preliminary Observation on Real-World-Scale Models}
\label{sec:app-real-world}
Here we provide a comparison table of real-world-scale U-Net/DiT diffusion models in \Cref{tab:app-naive}.
Since there are many confounders, such as model architectures, parameter sizes, or training recipes, we do not aim to draw very detailed conclusions from it.
However, we can observe that most DiT models yield weaker features than U-Net models, and this gap extends across many different tasks, not restricted to semantic correspondence.

\begin{table*}[t]
\centering
\caption{Comparison of diffusion features extracted from prevalent large-scale text-to-image models on all six settings in \Cref{tab:setting}. Best and second are marked as \textorange{\textbf{bold}} and \textblue{\underline{underlined}}.}
\label{tab:app-naive}
\begin{tabular}{@{}lccccccc@{}}
\toprule
Model  & \#Params      & SPair (a) & SPair (b) & SPair (c) & Horse-21 & CityScapes & Depth \\ \midrule
SD1.5  &1B      & 62.79      & 77.78      & 86.04      & 57.90 & 64.10  & 0.86      \\
SD2.1  &1B      & 65.20      & 80.00      & 86.73      & 58.21   & 65.23 & 0.88   \\
SDXL   &3.5B      & 63.14      &81.72      & 88.68      & \textorange{\textbf{63.20}}   & 74.47 & 0.83    \\
PGV2   &3.5B      & 65.66      & \resultsecond{83.44}      & \textorange{\textbf{92.31}}      & 61.73  & \resultsecond{74.74} & \resultsecond{0.93}  \\ \midrule
PixArt-Sigma &0.6B & 60.23      & 72.77      & 81.72      & 55.28   & 26.17 & 0.77    \\
HunyuanDiT & 1.5B   & \textblue{\underline{67.88}}      & 80.50      & \resultsecond{90.02}      & \resultsecond{62.93} & \textorange{\textbf{77.50}}   & \textorange{\textbf{0.94}} \\
SD3.5M &2.5B      & 56.75      & 68.72      & 65.93      & 51.77 & 59.38 & 0.79     \\
SD3.5L &8B      & 65.09      & 76.86      & 79.16      & 51.68 & 55.13 & 0.78     \\
Flux   &12B      & \textorange{\textbf{72.62}}      & \textorange{\textbf{85.93}}      & 89.67      & 61.58  & 74.32 & 0.92  \\  \bottomrule
\end{tabular}
\end{table*}

\subsection{Effect of Backbone Prediction Type}
\label{sec:app-evx}

The diffusion backbone can be trained to predict three different targets: $\epsilon$, $v$, and $x$~\cite{jit}.
$\epsilon$-prediction is the common choice for DDPM, while $v$-prediction has become more popular with the shift into flow matching.
In comparison, $x$-prediction is only preliminarily explored~\cite{jit}.
In this subsection, we will show that the prediction target can affect the effect of adding LSCs to some extent.
However, the prediction target is relatively superficial, while the peak position of the quality-layer curve is more essential.

In the main manuscript, we are using an $\epsilon$-prediction code to train our models.
Here, we switch to a $v$-prediction code and repeat the experiment in \Cref{tab:lsc-exp}.
This code is based on SiT~\cite{sit}, which is largely derived from the codebase of \cite{Peebles23dit}, using the same model architecture but switched to flow matching.
Additionally, we propose a new LSC implementation as $x \leftarrow x + \gamma \cdot \text{LayerNorm}(\text{skip})$, with $\gamma$ initialized to 0, because of training stability concerns.
We will later denote this new implementation as the one with \textit{residual connection}.
The results are shown in \Cref{tab:app-v-pred}.
As for $x$-prediction models, we use the JiT~\cite{jit} code, where we also use the LSC implementation with a residual connection.
The results are shown in \Cref{tab:app-x-pred}.

\begin{table}[t]
\centering
\caption{Comparing $v$-prediction DiTs without and with LSCs.}
\label{tab:app-v-pred}
\begin{tabular}{@{}lccccccc@{}}
\toprule
Model   & FID            & SPair (a)      & SPair (b)      & SPair (c)     & Horse-21       & CityScapes     & Depth         \\ \midrule
w/o LSC & 22.00             & 33.65          & 48.79          & 60.19         & 29.73          & 50.54          & 0.74          \\
w/ LSC  & \resultbest{21.53} & \resultbest{34.91} & \resultbest{51.28} & \resultbest{60.30} & \resultbest{30.37} & \resultbest{50.65} & \resultbest{0.75} \\ \bottomrule
\end{tabular}
\end{table}

\begin{table}[t]
\caption{Comparing $x$-prediction DiTs without and with LSCs.}
\label{tab:app-x-pred}
\begin{tabular}{@{}lccccccc@{}}
\toprule
Model   & FID   & SPair (a) & SPair (b)      & SPair (c)      & Horse-21    & CityScapes & Depth      \\ \midrule
w/o LSC & 40.42 & 14.49     & \resultbest{26.38} & \resultbest{45.16} & \resultsecond{11.91} & 41.65      & \resultsecond{0.66} \\
w/ LSC     & \resultsecond{39.49}    & \resultbest{14.87} & \resultsecond{\resultbest{26.38}} & 43.40        & \resultbest{11.93} & \resultsecond{41.80}     & \resultbest{0.67} \\
skewed LSC & \resultbest{38.35} & \resultsecond{14.76}    & \resultsecond{25.92}          & \resultsecond{44.74} & 11.58          & \resultbest{42.21} & \resultbest{0.67} \\ \bottomrule
\end{tabular}
\end{table}

\begin{figure*}[t]
    \centering
    \includegraphics[width=\linewidth]{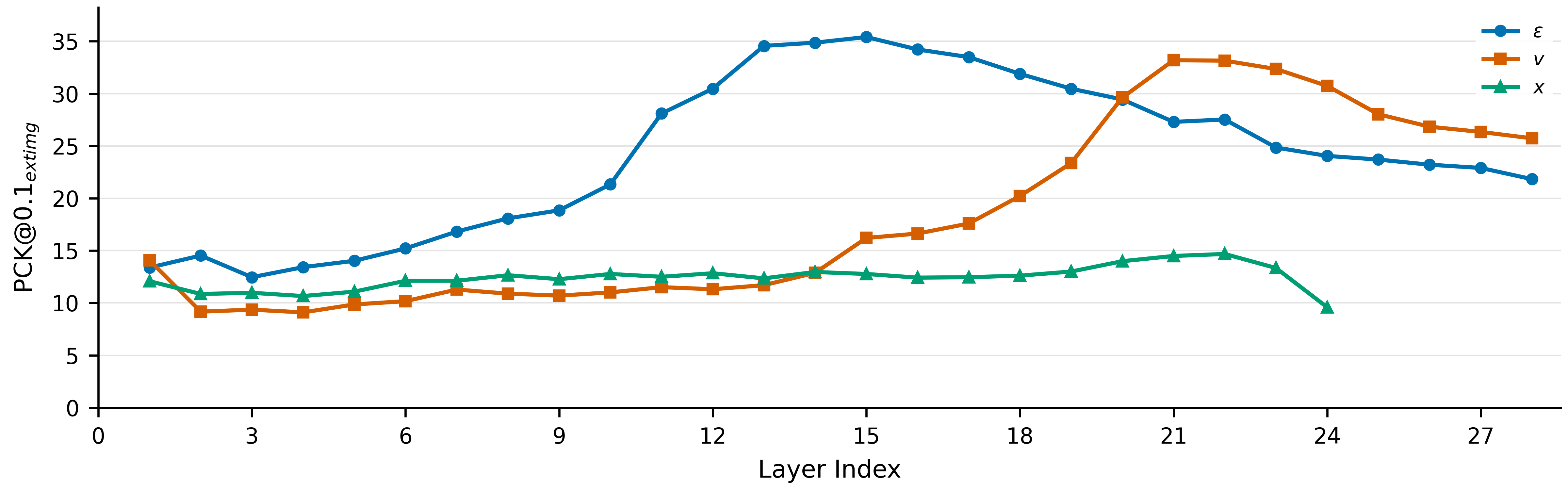}
    \caption{Quality-layer curves from three different prediction types.}
    \label{fig:prediction-type}
\end{figure*}

Before further discussion, we introduce two new concepts.
(i) The peak position of quality-layer curves.
To plot a quality-layer curve, we use SPair (a) performance as the y-axis and the layer index as the x-axis.
This curve peaks at a certain layer, which can be at the middle layer or not, determined by complex factors.
The quality-layer curves of models without LSCs are plotted in \Cref{fig:prediction-type}, showing a clear shift of peak position.
(ii) The topology of LSCs.
By default, the LSC connections are symmetric, but it is legal to design a topology that is not.
For instance, for a 28-layer DiT backbone, we could put the symmetry center between layer 10 and 11, instead of layer 14 and 15, creating a \textit{skewed} topology.
This effectively introduces a structural bias that tells the model to put its peak position at this skewed symmetry center.

From the results above, we can see that the feature quality is not significantly affected, but generation is overall better.
Furthermore, the skewed LSC topology is better than the symmetric one when the vanilla peak position is already skewed.
In addition, we have also tried to train SiT and JiT models using the same LSC structure (the one without residual connection) as used in the main manuscript.
This structure tends to bring more significant representation quality enhancement and force the quality-layer curve to align with the LSC topology, but does not guarantee generation improvement.
In general, we believe that whether and how the vanilla quality-layer curve skews is more essential, while the exact prediction target is relatively superficial.
Nevertheless, how this interaction actually works remains an open question.
For practical suggestions, we believe it is more beneficial if the LSC topology is designed to match the vanilla peak position of quality-layer curves.

\subsection{Comparing DiT-XL/2, U-ViT, and Skip-DiT}
\label{sec:app-pretrained}

To preliminarily support our hypothesis that LSCs contribute to better feature quality, we compare two open-source LSC-augmented DiT models from the literature:
\begin{enumerate}[label=(\roman*)]
    \item ImageNet 256x256 (U-ViT-H/2) of U-ViT~\cite{u-vit}.
    \item DiT-XL/2-skip for Skip-DiT~\cite{skip-dit}.
\end{enumerate}
As a baseline with no LSCs, we also include the DiT-XL/2 (256) model for DiT~\cite{Peebles23dit}.
We have confirmed that they share the same architecture hyperparameters except for LSCs.
However, it should be noted that their training hyperparameters, such as training steps and batch size, are likely to differ, but the final FID scores roughly align.

\begin{table}[t]
\centering
\caption{Comparing open-source DiT models with LSCs. Best results marked as \resultbest{bold}. The FID scores are taken from the official reports, and the discriminative capability is evaluated using the six settings in \Cref{tab:setting}.}
\label{tab:app-pretrained}
\begin{tabular}{@{}lccccccc@{}}
\toprule
Model    & FID           & SPair (a)      & SPair (b)      & SPair (c)      & Horse-21       & CityScapes     & Depth         \\ \midrule
DiT      & \resultbest{2.27} & 44.01          & 64.89          & \resultbest{77.62} & 39.29          & 53.04          & 0.76          \\
U-ViT    & 2.29          & \resultbest{53.53} & 66.69          & 75.02          & \resultbest{40.53} & \resultbest{53.97} & \resultbest{0.84} \\
Skip-DiT & 2.29          & 41.91          & \resultbest{68.29} & 73.61          & 39.08          & 52.53          & 0.73          \\ \bottomrule
\end{tabular}
\end{table}

From \Cref{tab:app-pretrained}, we can make the following observations:
\begin{enumerate}[label=(\roman*)]
    \item Comparing U-ViT and DiT, U-ViT achieves overall better representation quality under a similar generation capability to DiT.
    We can attribute this gain to the LSC structure with some confidence, since the difference between the two models is LSCs \textit{and} the training recipe.
    This motivates us to hypothesize that LSCs contribute to diffusion feature quality, but more evidence is still required to further exclude the effect of the training recipe.
    \item Skip-DiT does not yield comparable results under certain settings.
    Note that, as argued as a main advantage, Skip-DiT has been through much fewer training steps before reaching a 2.29 FID score, which might cause its inner representations to be weaker.
    In general, this suggests the necessity of a from-scratch training experiment in a better-controlled environment to provide more concrete evidence.
    \item If the diffusion backbone itself is fine-tuned during the discriminative benchmark (SPair (c)), the gap between non-LSC- and LSC-DiT models can become narrower and even reversed. This may be because the backbone is supervised entirely using the discriminative training signal. This effectively discards the latter half of the backbone and no longer requires the final noise prediction capability. Hence, even for a DiT backbone with no LSCs, it can safely lower the diffusion noise intensity in middle layers, reaching a similar behavior to LSC-DiT models.
\end{enumerate}

\subsection{Mutual Information Probing Results}
\label{sec:app-mi}

\begin{figure*}[t]
    \centering
    \includegraphics[width=1\linewidth]{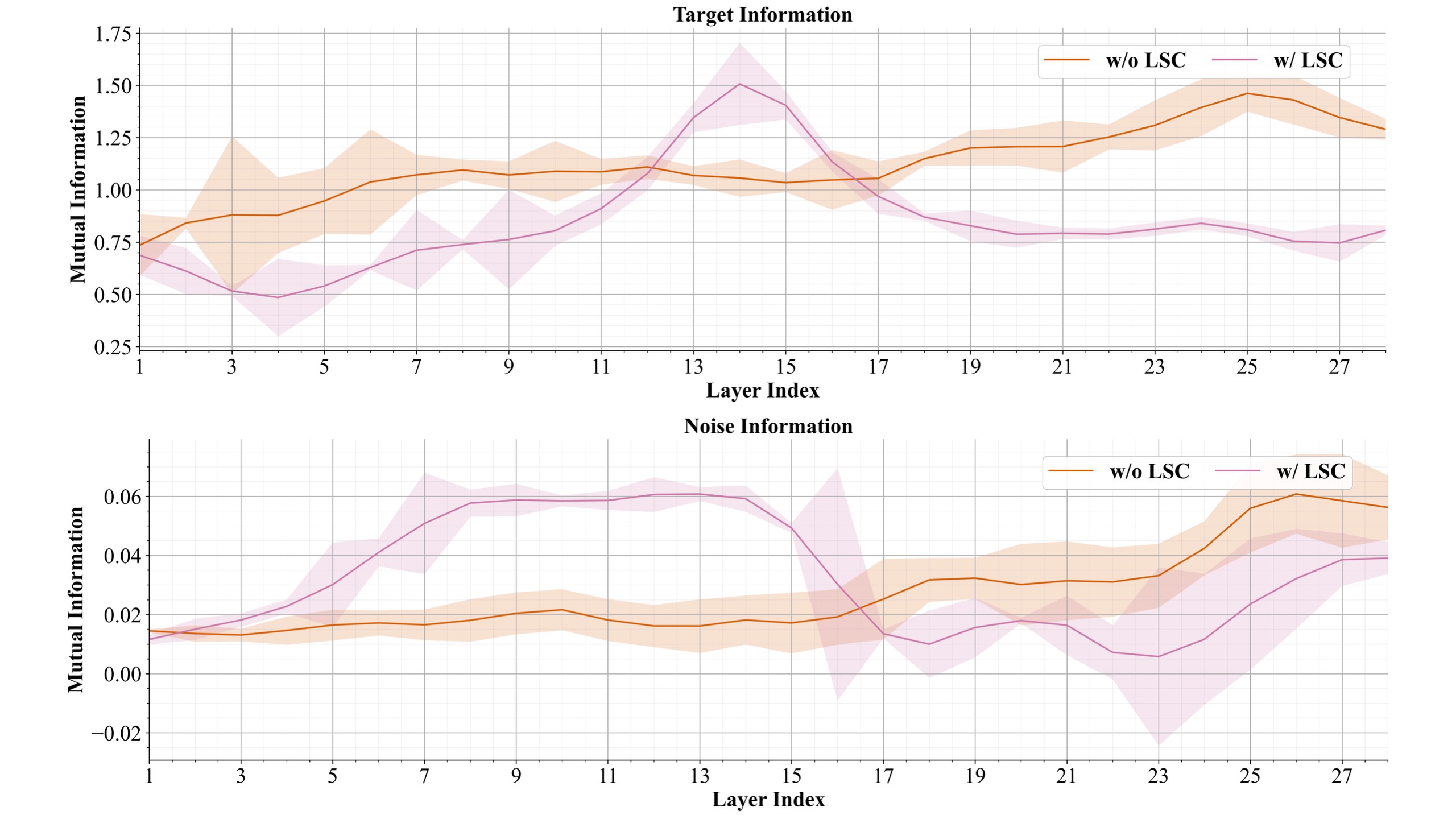}
    \caption{Mutual information between each layer's activation and noise/target information, measured using MINE, under the second setting.}
    \label{fig:app-mi-2}
\end{figure*}

Considering that MINE can be sensitive, we repeat the experiment under two settings: (MINE hidden channel as 512, projection head channel as 256, training steps as 600) and (MINE hidden channel as 1024, projection head channel as 512, training steps as 1000).
Moreover, for each setting, we repeat this experiment under three different seeds, each time re-extracting all features.
The result from the first setting is reported in \Cref{fig:mech-mi}, and the results from the second setting are reported in \Cref{fig:app-mi-2}.
We can draw the same conclusions from the two experiments that our noise bypassing hypothesis could be supported.

Additionally, we also report the implementation details of the MINE probing.
The input to the system is a 2D feature map of multiple channels.
It is first projected into the MINE estimator dimension with a $1\times 1$ convolutional layer.
Then, we use average pooling to shrink it into a global vector per sample.

\subsection{Two Corollaries of Noise Bypassing Hypothesis}
\subsubsection{Duo-Peak and Single-Peak Quality-Layer Curves}
\begin{figure*}[t]
    \centering
    \includegraphics[width=.9\linewidth]{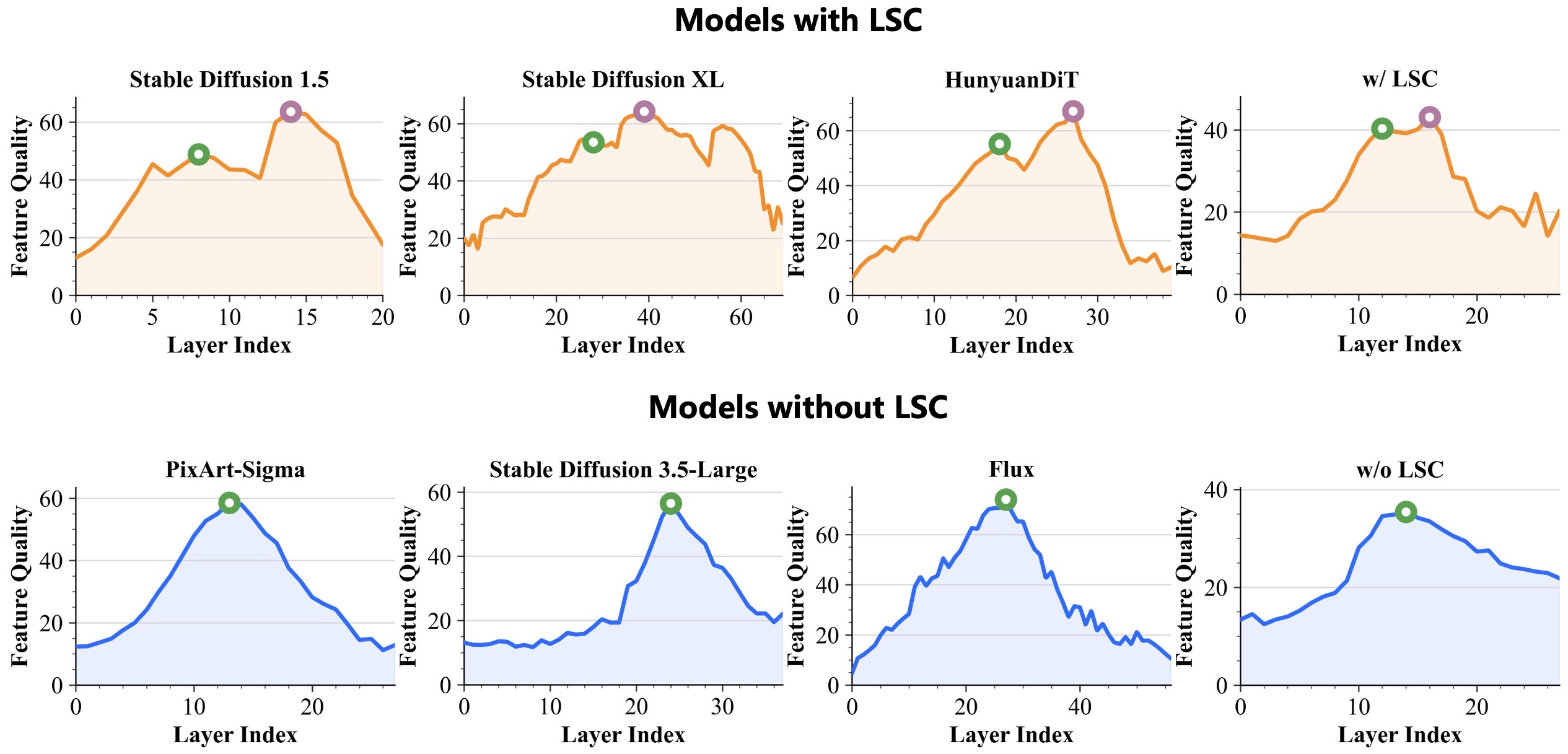}
    \caption{Quality-layer curves (x-axis for layer depth and y-axis for feature quality) of different diffusion backbones. Green dots mark the first peaks, and purple dots mark the second peaks. Quality is evaluated under the \textit{Spair (a)} setting of \Cref{tab:setting}.}
    \label{fig:duo-peak}
\end{figure*}

If we plot a figure with the layer index of a diffusion backbone as the x-axis and the understanding capability of the feature extracted from this layer (evaluated under a certain discriminative task) as the y-axis, we obtain a quality-layer curve.
We observe that diffusion models without LSCs tend to manifest a quality-layer curve with a single peak, and models with LSCs tend to manifest a duo-peak curve, as shown in \Cref{fig:duo-peak}.

Next, we provide a hypothesized explanation for this observation.
Based on the three-section division of a diffusion backbone, features through the noise suppression section gradually grow in quality as they go into deeper layers in this section, as the noises are gradually suppressed, which forms the rising slope before the first peak.
Then, in the noise prediction section, noise increases, causing the final descending slope.
Importantly, the behavior of the target reconstruction section differs based on whether there are LSCs.
According to another finding in \cite{my-SDXL}, when trying to reconstruct an internal representation for the target image, activations capturing mostly abstract and coarse-grained information, such as positional and overall shape, are first formed, after which activations become more and more detailed.
Notably, activations containing abstract and coarse-grained information, when used as a diffusion feature, tend to be inferior to the other features.
Therefore, when there are LSCs, there will be a slight drop between the two peaks.
In comparison, with no LSCs, the activations in the target reconstruction section would be relatively noisy, and the noises overrun the aforementioned detailed change, showing only one peak in the curve.

\subsubsection{Upper and Lower LSCs}
According to the noise bypassing hypothesis, LSCs in the upper part of the ``U'' shape have access to stronger noises, so they should be able to transport stronger noise information, making them more important compared to lower LSCs.
Thus, if we keep only half of the LSCs, keeping the upper half should yield a better result than keeping only the lower half.

\subsubsection{Validation for the Two Corollaries}
\begin{figure}[t]
    \centering
    \includegraphics[width=.6\linewidth]{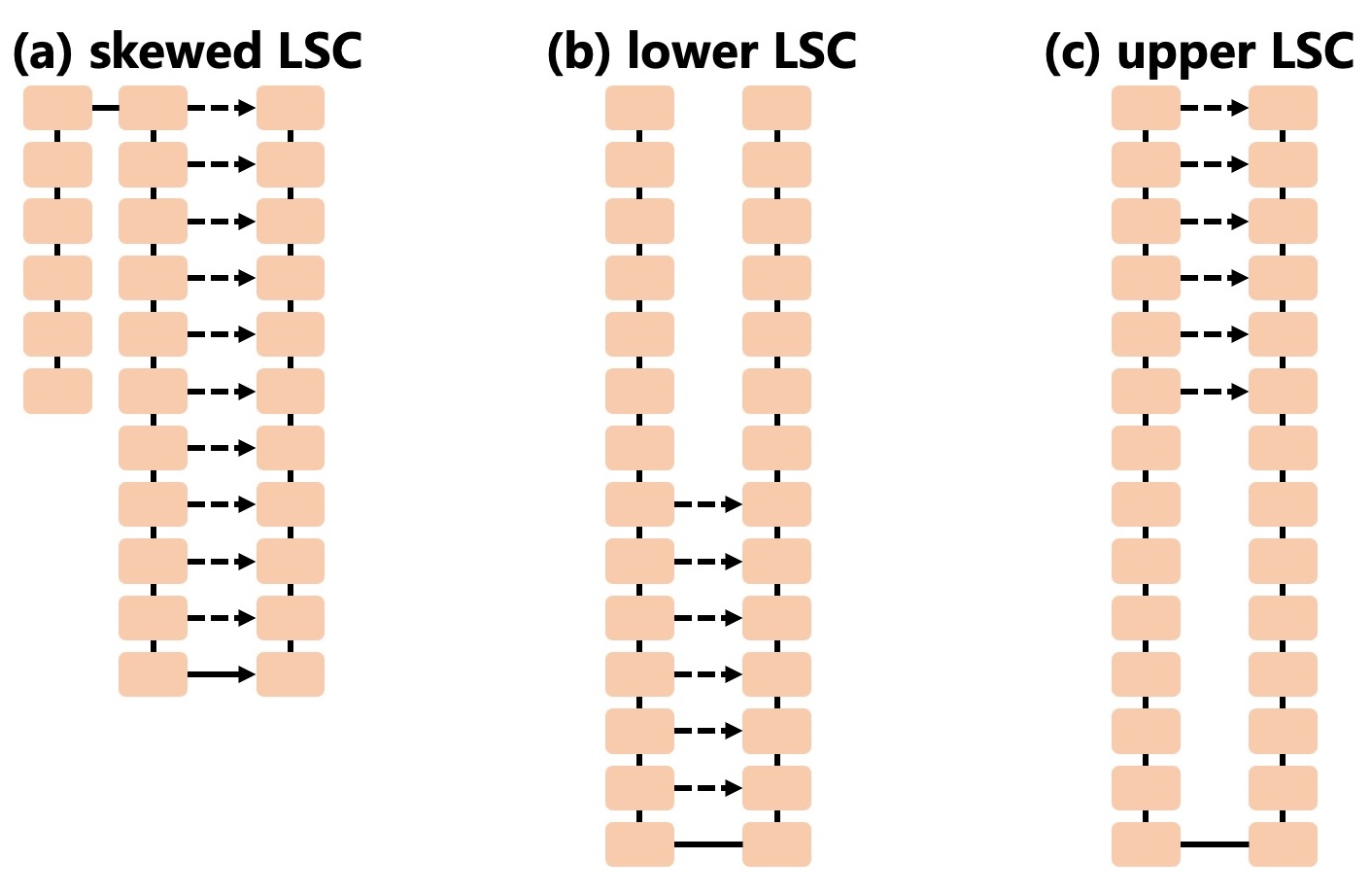}
    \caption{Different LSC designs to be tested in experiments. We use dotted lines to indicate LSCs. In this figure, the number of layers and position of LSCs \textit{exactly} reflect the architecture of the modified DiT models.}
    \label{fig:all-lsc}
\end{figure}

To validate the two corollaries, we conduct an additional controlled experiment.
In this experiment, we design several special LSC structures, as depicted in \Cref{fig:all-lsc}:
\begin{itemize}
    \item (a) \textit{skewed LSC}: The U-shape is skewed to one direction, possibly affecting the position of the second peak.
    \item (b) \textit{upper LSC}: To test the relative contribution of the upper LSCs, we actually keep only the upper ones.
    \item (c) \textit{lower LSC}: As a symmetric setting, only the lower LSCs are kept.
\end{itemize}

\begin{figure}[t]
    \centering
    \includegraphics[width=1\linewidth]{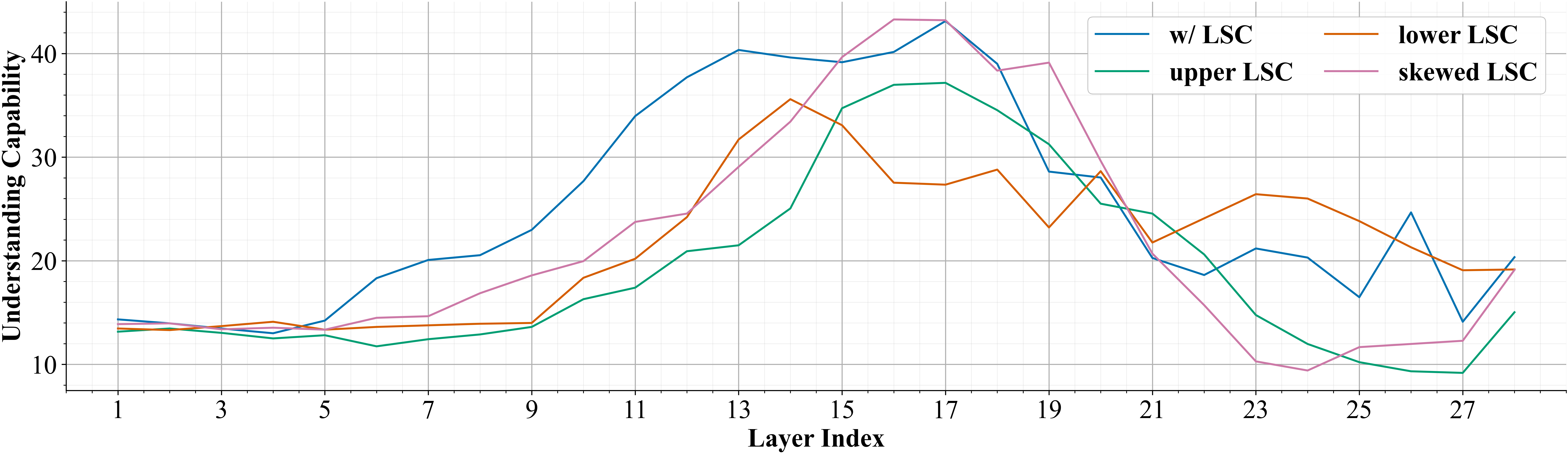}
    \caption{Quality-layer curves under three LSC variants. The quality is evaluated using \textit{SPair (a)} setting.}
    \label{fig:main-results}
\end{figure}

The quality-layer curves are visualized in \Cref{fig:main-results}, and the quantitative results are shown in \Cref{tab:mech-exp}.
\begin{enumerate}[label=(\roman*)]
    \item First, we can observe from \Cref{fig:main-results} that the skewed LSC design indeed makes the second peak skewed as well, thus validating the causal relation between LSCs and the duo-peak observation.
    \item Then, we can observe from \Cref{tab:mech-exp} that the upper LSCs are indeed more important than lower LSCs: if we only keep part of the LSCs, keeping the upper ones leads to stronger downstream performance than keeping the lower ones. Nevertheless, the lower LSCs still matter, as both variants are worse than the previous complete LSC setting in \Cref{tab:lsc-exp}.
\end{enumerate}
In general, this experiment validates the two corollaries of the noise bypassing hypothesis, indirectly strengthening the evidence supporting this hypothesis.

\begin{table}[t]
\centering
\caption{Discriminative performance evaluated on all six settings the same as \Cref{tab:lsc-exp}. Best/second marked as \textorange{\textbf{bold}} and \textblue{\underline{underlined}}.}
\label{tab:mech-exp}
\begin{tabular}{@{}lcccccc@{}}
\toprule
LSC Setting      & SPair (a) & SPair (b) & SPair (c) & Horse-21 & CityScapes & Depth \\ \midrule
skewed LSC & \textorange{\textbf{42.33}}      & \textorange{\textbf{58.93}}      & \textorange{\textbf{66.76}}      & \textorange{\textbf{37.08}}      & \textorange{\textbf{51.16}}      & \textorange{\textbf{0.74}}       \\
upper LSC  & \resultsecond{36.86}      & \resultsecond{52.69}      & \resultsecond{61.56}      & \resultsecond{31.91}      & \resultsecond{50.11}      & \textblue{\underline{0.71}}       \\
lower LSC  & 35.60      & 52.54      & 61.18      & 28.72      & 49.55      & 0.70       \\ \bottomrule
\end{tabular}
\end{table}

\subsection{Attempts to Add LSCs to Pre-Trained DiTs}
\label{sec:add-lsc}
\subsubsection{Method}
We have tried several different methods to add LSCs to pre-trained DiTs, basically divided into two-stage and single-stage training types.

\noindent \textbf{Two-Stage Method}.
Following the adaptation method in Skip-DiT~\cite{skip-dit}, we divide the training into two stages.
At the first stage, we freeze the backbone and only train LSC connectors until they converge, indicated by roughly recognizable generation results.
Then, we unfreeze the backbone and train the entire model altogether.
Ideally, during the second stage, the generative capability would quickly restore and improve, but we did not observe such behavior.
Moreover, we notice that the features in the upsampling stage become very noisy.
Therefore, we conclude that the two-stage fine-tuning method might be applicable only to certain diffusion models.
We hypothesize that the first-stage training introduces too much degradation to the model, which is hard to correct in the second stage.
To tackle the drawback of this method, we further propose a new single-stage method.

\noindent \textbf{Single-Stage Method}.
Directly adding a vanilla LSC structure into a pre-trained DiT backbone will damage its original behavior and make single-stage fine-tuning hard, as we need to restore the generative capability from a low point.
To tackle this, we propose the zero-init LSC design, which aims to make no impact on model behavior at initialization.
In the zero-init LSC design, we put a residual connection in the LSC:
\begin{equation}
    x = x + \text{Linear}(\text{Norm}(skip))
\end{equation}
where Norm is layer normalization, and Linear is a linear layer.
The weight and bias of this linear layer are both initialized to zero, making the residual connection degenerate to an identity projection at initialization.
Thanks to zero-init LSCs, we can safely fine-tune the entire model from the beginning using the standard diffusion training objective:
\begin{equation}
    \mathcal{L}=\mathbb{E}_{(x_0,c)\sim\mathcal{D},x_1,t}||v'_\theta(z_t,t,c)-v_t||^2
\end{equation}
where $v'$ is the DiT backbone augmented with LSCs, $v'_\theta(z_t,t,c)$ is the predicted velocity, and $v_t$ is the ground truth.

Note that the initial state of the zero-initialized model is already at a local optimum, so we need to add some additional regularization to force the model out of it.
Specifically, we extract features from the main activation residual flow in the noise suppression section and the target reconstruction section.
First, we force the features from the target reconstruction to be clean of noises.
Then, we encourage the features from the noise suppression section to be noisier.
In this way, we hope the model can learn to shortcut the noise information through the newly added LSCs.
We find that noises can be easily suppressed by a Gaussian blur, which creates a ground truth for this regularization term:
\begin{equation}
    \mathcal{L}_{noise} = \text{MSE}(feat, \text{sg}(\text{GaussianBlur}(feat)))
\end{equation}
where MSE is the mean squared error loss, $feat$ is the extracted feature, GaussianBlur is the Gaussian blur operation, and sg denotes the stop-gradient operator.
We find that a weight of $0.001$ / $-0.001$ is generally stable for section 2 / section 1, where the negative weight should be combined with gradient clipping to alleviate instability.

\subsubsection{Training Strategy}
We fine-tune PixArt-Sigma with the single-stage method to evaluate the viability.
To construct the training data, we adopt a dataset titled DiffusionDB~\cite{wangDiffusionDBLargescalePrompt2022}.
This dataset contains prompts and the corresponding images used by real diffusion users.
We only use its prompt part, letting the original PixArt-Sigma generate images to obtain image-prompt pairs.
In this way, we hope to retain the original generation capability with the fine-tuned model.

\subsubsection{Results}
\begin{figure}[t]
    \centering
    \includegraphics[width=1\linewidth]{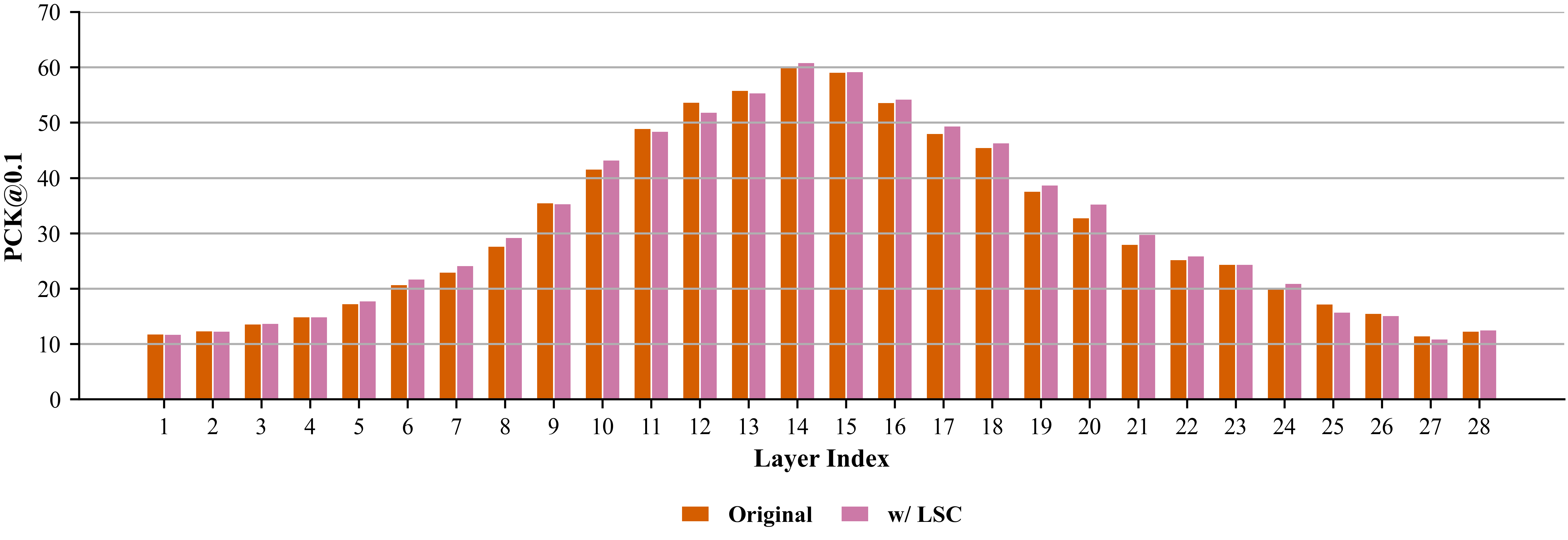}
    \caption{Per-layer feature quality from the original model and the model fine-tuned with LSC inserted. The quality is evaluated using $\text{PCK@0.1}_{\text{img}}\uparrow$.}
    \label{fig:app-lsc}
\end{figure}
\begin{figure}[t]
    \centering
    \includegraphics[width=1\linewidth]{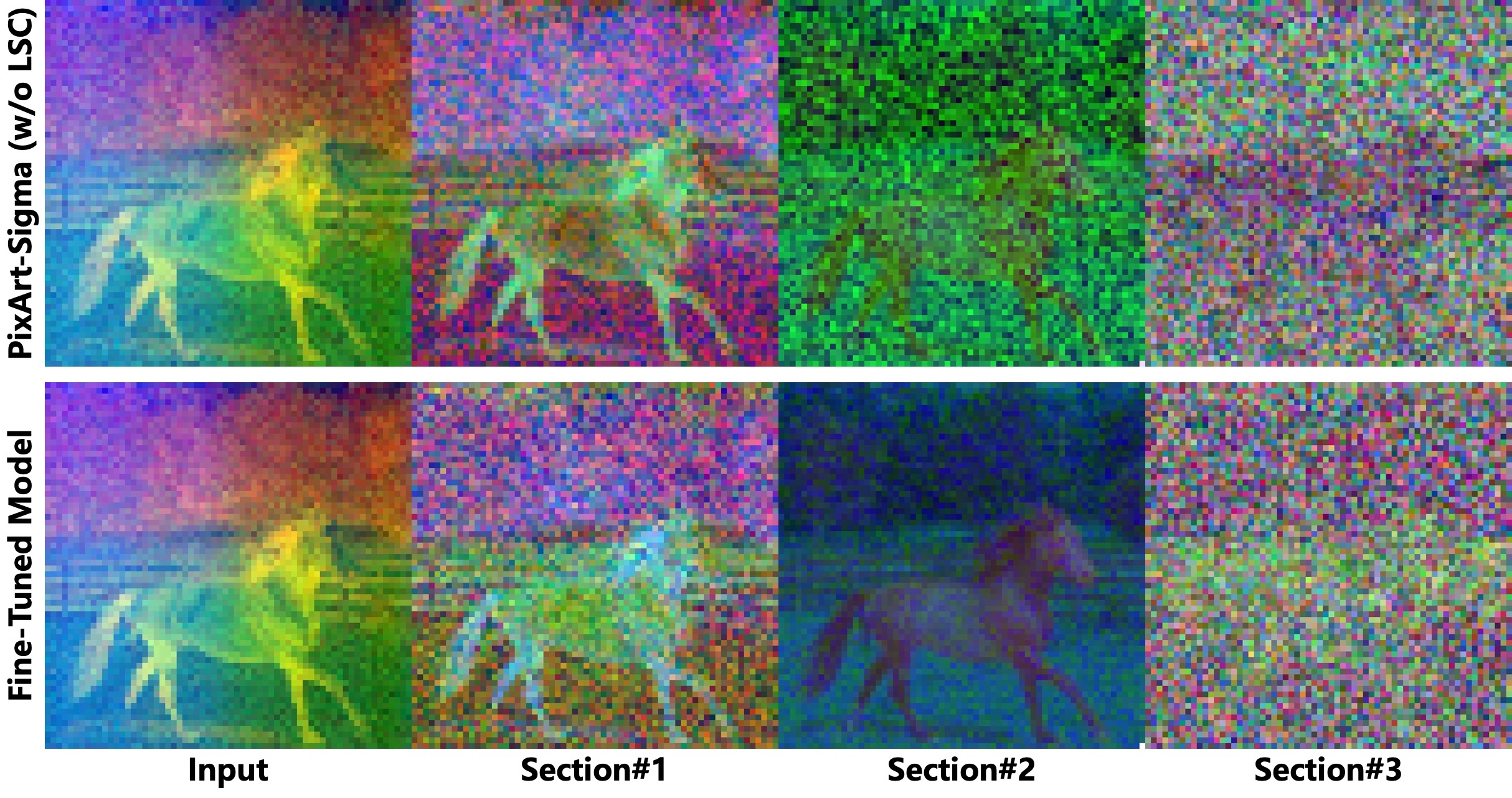}
    \caption{Feature PCA visualization of the original PixArt-Sigma model and the fine-tuned version with LSCs.}
    \label{fig:app-lsc-vis}
\end{figure}
We compare the feature quality, evaluated using \textit{Spair (a)} settings in \Cref{tab:setting}, of both the original and fine-tuned models in \Cref{fig:app-lsc}.
We can notice that:
\begin{enumerate}[label=(\roman*)]
    \item Feature quality increases, but only slightly.
    \item The overall quality-layer curve is roughly the same as before, showing that the model's inner behavior is not fundamentally changed to manifest the duo-peak curve.
\end{enumerate}
In addition, we provide some feature PCA visualization in \Cref{fig:app-lsc-vis}.
We can observe that the additional noise regularization term can actually toggle the noise strength in different parts of the backbone to some extent, but the result is not satisfying enough.
For example, in the target reconstruction section, the features contain less noises than the original PixArt-Sigma model, but still more than HunyuanDiT.

In summary, we believe that the strong path dependence of DiTs’ learned representations makes it tough to fundamentally change the behavior of pre-trained DiT models.
To be specific, the backbone model has already learnt to transport both noise and target information in the same activations, which might be the optimal behavior when there are no LSCs.
Such an ability might conflict with the ability to separately transport the information of two aspects in different paths, making fine-tuning particularly difficult.
Therefore, fundamentally changing how the backbone functions via fine-tuning might require as much training as training from scratch, or even more.
Currently, we believe that, to effectively fuse LSCs into a DiT, it is likely that the model needs training from the start, and direct retrofitting is likely to be hard in nature.
Nevertheless, it is not impossible that, with a better fine-tuning strategy, we could really add LSCs to a pre-trained DiT model.

\subsection{Computational Cost of LSC}
Computational cost is essential nowadays~\cite{xiao2026green}.
We provide the parameter size of different LSC settings in \Cref{tab:size}.
We can observe that, parameter-wise, LSCs only introduce a slight overhead, compared to their huge bonus to both feature quality and generation.
Additionally, for more practical DiT architectures with more complex designs, this overhead will become even smaller.
In our experiments, the running time of different LSC designs is almost identical.

However, it is possible that LSCs might conflict with some acceleration techniques based on parallelism by introducing additional shortcut computation.
This may also be one of the reasons why LSCs are not commonly used in today's DiTs.
Nevertheless, there are also methods that utilize LSCs to accelerate diffusion inference, such as caching~\cite{skip-dit}.
By adopting LSCs in future DiT designs, such methods would be enabled for application, which might make LSCs in DiT more appealing efficiency-wise.




\begin{table*}[t]
\centering
\caption{Parameter sizes of different LSC designs, compared to the original \textit{w/o LSC}.}
\label{tab:size}
\begin{tabular}{@{}lcc@{}}
\toprule
LSC Type         & Parameter Size & Percentage w.r.t. w/o LSC \\ \midrule
w/o LSC          & 675,129,632    & 100.0\%                   \\
w/ LSC     & 709,709,216    & 105.1\%                   \\
upper LSC  & 691,089,440    & 102.4\%                   \\
lower LSC  & 691,089,440    & 102.4\%                   \\
skewed LSC & 701,729,312    & 103.9\%                   \\ \bottomrule
\end{tabular}
\end{table*}

\subsection{Exact Feature Extraction Specification for DiTs}
\label{sec:app-config}
In this section, we specify the exact optimal feature extraction spots and timesteps for the DiTs evaluated in \Cref{tab:app-naive}.
The extraction spots and timesteps are established following the pipeline in \Cref{sec:pre}, and once the optimal choice is decided, it is reused across all experiments.
The layer index starts at 0, and the timestep is given based on a max timestep of $T=1000$.

PixArt-Sigma ($t=100$):
\begin{enumerate}[label=(\roman*)]
    \item Cross-attention query at layer 13.
    \item Output of self-attention at layer 13.
    \item Output of the 2nd normalization at layer 14.
\end{enumerate}
HunyuanDiT ($t=50$):
\begin{enumerate}[label=(\roman*)]
    \item The normalized cross-attention query at layer 27.
    \item Self-attention query at layer 27.
    \item Output of the 3rd normalization at layer 26.
\end{enumerate}
SD3.5M ($t=300$):
\begin{enumerate}[label=(\roman*)]
    \item Output of the 1st attention at layer 9.
    \item Output of the 2nd normalization at layer 9.
    \item Query of the 1st attention at layer 11.
\end{enumerate}
SD3.5L ($t=400$):
\begin{enumerate}[label=(\roman*)]
    \item Output of the 2nd normalization at layer 23.
    \item Output of the 1st attention at layer 23.
    \item Value of the 1st attention at layer 24.
\end{enumerate}
Flux ($t=200$):
\begin{enumerate}[label=(\roman*)]
    \item Attention output at layer 27.
    \item Attention key at layer 29.
    \item Attention output at layer 25.
\end{enumerate}
\textit{w/o LSC} ($t=50$):
\begin{enumerate}[label=(\roman*)]
    \item Output activation at layer 13.
    \item Output activation at layer 14.
    \item Output activation at layer 15.
\end{enumerate}
\textit{w/ LSC} ($t=50$):
\begin{enumerate}[label=(\roman*)]
    \item Output activation at layer 14.
    \item Output activation at layer 15.
    \item Output activation at layer 16.
\end{enumerate}
\textit{upper LSC} ($t=50$):
\begin{enumerate}[label=(\roman*)]
    \item Output activation at layer 15.
    \item Output activation at layer 16.
    \item Output activation at layer 17.
\end{enumerate}
\textit{lower LSC} ($t=50$):
\begin{enumerate}[label=(\roman*)]
    \item Output activation at layer 13.
    \item Output activation at layer 14.
    \item Output activation at layer 15.
\end{enumerate}
\textit{skewed LSC} ($t=50$):
\begin{enumerate}[label=(\roman*)]
    \item Output activation at layer 15.
    \item Output activation at layer 16.
    \item Output activation at layer 17.
\end{enumerate}
In addition, the U-Net features are extracted at the optimal spots specified in \cite{my-SDXL}.

Note that the specification above is not applicable to quality-layer curves.
For quality-layer curves, we extract features from similar spots of each layer at the optimal timestep as stated above.
For example, we can extract the basic block output of each layer to plot the quality-layer curves.

\subsection{Experimental Details}
\textbf{Semantic Correspondence}~\cite{hao2020brief}:
This task aims to predict the corresponding keypoint in the target image given a keypoint in the source image, where the two keypoints should be semantically similar.
The dataset used is SPair-71k~\cite{min2019spair}.
$\text{PCK@0.1}_{\text{img}}(\uparrow)$ and $\text{PCK@0.1}_{\text{bbox}}(\uparrow)$ are commonly used metrics, following the widely adopted protocol reported in \cite{min2019spair}. These two metrics measure the percentage of correctly predicted keypoints, where a predicted keypoint is considered to be correct if it lies within the neighborhood of the corresponding annotation with a radius of $0.1 \times max(h, w)$. For $\text{PCK@0.1}_{\text{img}}$/$\text{PCK@0.1}_{\text{bbox}}$, $h, w$ denote the dimensions of the entire image/object bounding box, respectively.
In Setting\#2, we train a small downstream network, which is a two-layer convolutional network.
In the other two settings for this task, we directly perform nearest neighbor computation, so no downstream model is required.
If training is required, we use a learning rate of 1e-4 for the downstream network and 1e-5 for the diffusion backbone, and the optimizer is AdamW for smaller models and AdamW8bit for larger diffusion models (SD3.5L and Flux).
The prompt used for feature extraction is ``a highly realistic photo that may contain an aeroplane, a bicycle, a bird, a boat, a bottle, a bus, a car, a cat, a chair, a cow, a dog, a horse, a motorbike, a person, a plant within a pot, a sheep, a train, or a tv monitor''.

\textbf{Semantic Segmentation}:
This task aims to classify pixels into semantic categories~\cite{yang2026gala}.
The dataset used is CityScapes~\cite{Cordts16city}.
We use mIoU metric, which is the mean over the IoU performance across all semantic classes \cite{hao2020brief}. For each image, IoU (Intersection over Union, $\uparrow$) is defined by \#(overlapped pixels between the prediction and the ground truth) / \#(union pixels of them).
In this setting, we train a relatively large downstream network, which is UPerNet~\cite{xiao2018unified}.
We use a learning rate of 0.00016 and AdamW optimizer.
The prompt used for feature extraction is ``An urban street scene with multiple lanes, various buildings, traffic lights, cars in the lanes, and pedestrians, highly realistic.''.

\textbf{Label-Scarce Semantic Segmentation}~\cite{baranchuk2021label}:
This is a special setting for semantic segmentation, where the labeled training data is very scarce (\~30 sample-annotation pairs).
The dataset used is collected in \cite{baranchuk2021label}, and we experiment on its Horse-21 subset, the data of which is sourced from LSUN~\cite{yu15lsun}.
In this setting, we train 10 pixel-classification MLP models and aggregate their predictions by choosing the majority.
The learning rate is 0.001, and the optimizer is Adam.
The prompt used for feature extraction is ``a horse in the field running, high quality, photorealistic''.

\textbf{Depth Estimation}~\cite{DBLP:conf/iclr/BochkovskiyDGSZ25}:
This task aims to regress the depth value, \textit{i.e.}, the distance of each pixel from the camera.
The dataset used is NYU Depth v2~\cite{nyudepthv2}.
We use d1 as the evaluation metric.
It computes the percentage of correctly predicted pixels, where a pixel is counted as correct if the difference between its value and the ground truth value does not exceed 25\%.
In this setting, we add a downstream model consisting of two regular convolutional layers and two 1$\times$1 convolution layers, training this model along with the diffusion backbone itself.
The diffusion backbone is trained with a 10-times smaller learning rate than the downstream network (1e-4/1e-5), and the optimizer is AdamW8bit.
The prompt used for feature extraction is ``a photo, focusing on the depth of the scene''.

\subsection{Computation Resources}
\label{sec:computation-resource}
To train the DiT models under different settings from scratch, we use $2\times$A100 GPUs for each experiment, which takes around five days per training run.
For the other experiments, such as evaluating these DiT models or the real-world-scale experiment, we only need a single 3090 or L20 GPU per experiment.
Whether we use 3090 or L20 depends on the parameter size of the pre-trained DiT model and the number of trainable parameters in the backbone.
Nevertheless, we need to point out that we have multiple cards to run multiple experiments in parallel, without which this study would be somehow hindered.

\subsection{Licenses}
\label{sec:app-license}
\begin{itemize}[leftmargin=10pt]
    \item SPair-71k: Available at \url{https://cvlab.postech.ac.kr/research/SPair-71k/}.
    \item Label-scarce segmentation datasets (sourced from LSUN): Available at \url{https://github.com/fyu/lsun}.
    \item CityScapes: Custom, at \url{https://www.cityscapes-dataset.com/license/}.
    \item NYU Depth v2: Available at \url{https://cs.nyu.edu/~fergus/datasets/nyu_depth_v2.html}.
    \item Stable Diffusion v1.5: The CreativeML OpenRAIL M license, available at \url{https://huggingface.co/stable-diffusion-v1-5/stable-diffusion-v1-5}.
    \item Stable Diffusion v2.1: CreativeML Open RAIL++-M License, available at \url{https://huggingface.co/Manojb/stable-diffusion-2-1-base}.
    \item Stable Diffusion XL: CreativeML Open RAIL++-M License, available at \url{https://huggingface.co/stabilityai/stable-diffusion-xl-base-1.0}.
    \item Playground v2: Playground v2 Community License, available at \url{https://huggingface.co/playgroundai/playground-v2-1024px-aesthetic}.
    \item PixArt-Sigma: CreativeML Open RAIL++-M License, available at \url{https://huggingface.co/PixArt-alpha/PixArt-Sigma-XL-2-1024-MS}.
    \item Stable Diffusion v3.5: the Stability Community License, available at \url{https://huggingface.co/stabilityai/stable-diffusion-3.5-medium} and \url{https://huggingface.co/stabilityai/stable-diffusion-3.5-large}.
    \item Flux: flux-1-dev-non-commercial-license, available at \url{https://huggingface.co/black-forest-labs/FLUX.1-dev}.
    \item HunyuanDiT: tencent-hunyuan-community, available at \url{https://huggingface.co/Tencent-Hunyuan/HunyuanDiT}.
    \item ImageNet: Custom terms of access (non-commercial research and educational use only), available at \url{https://image-net.org/}.
    \item DiffusionDB: CC0 1.0, available at \url{https://huggingface.co/datasets/poloclub/diffusiondb}.
    \item DiT: CC-BY-NC 4.0, available at \url{https://github.com/facebookresearch/DiT}.
    \item SiT: MIT License, available at \url{https://github.com/willisma/SiT}.
    \item JiT: MIT License, available at \url{https://github.com/LTH14/JiT}.
    \item U-ViT: MIT License, available at \url{https://github.com/baofff/U-ViT}.
    \item Skip-DiT: Apache-2.0 (code, available at \url{https://github.com/OpenSparseLLMs/Skip-DiT}); CC-BY-NC 4.0 (model weights, available at \url{https://huggingface.co/GuanjieChen/Skip-DiT}).
\end{itemize}


\newpage
\section*{NeurIPS Paper Checklist}

\begin{enumerate}

\item {\bf Claims}
    \item[] Question: Do the main claims made in the abstract and introduction accurately reflect the paper's contributions and scope?
    \item[] Answer: \answerYes{} 
    \item[] Justification: We tried our best to make the abstract and introduction reflect the contributions and scope of this paper.

\item {\bf Limitations}
    \item[] Question: Does the paper discuss the limitations of the work performed by the authors?
    \item[] Answer: \answerYes{} 
    \item[] Justification: We discuss the limitations mainly in the conclusion section (\Cref{sec:conclusion}) and explicitly add the phrasing ``limitation'' in the section title.

\item {\bf Theory assumptions and proofs}
    \item[] Question: For each theoretical result, does the paper provide the full set of assumptions and a complete (and correct) proof?
    \item[] Answer: \answerNA{} 
    \item[] Justification: Although, in our own opinion, this work is more like a foundational paper than a method contribution, it is more empirical than theoretical.

    \item {\bf Experimental result reproducibility}
    \item[] Question: Does the paper fully disclose all the information needed to reproduce the main experimental results of the paper to the extent that it affects the main claims and/or conclusions of the paper (regardless of whether the code and data are provided or not)?
    \item[] Answer: \answerYes{} 
    \item[] Justification: We have included many details. If there should be any missing information, it can be easily clarified with the code release.

\item {\bf Open access to data and code}
    \item[] Question: Does the paper provide open access to the data and code, with sufficient instructions to faithfully reproduce the main experimental results, as described in supplemental material?
    \item[] Answer: \answerNo{} 
    \item[] Justification: We plan to release the code if this work is accepted.

\item {\bf Experimental setting/details}
    \item[] Question: Does the paper specify all the training and test details (e.g., data splits, hyperparameters, how they were chosen, type of optimizer) necessary to understand the results?
    \item[] Answer: \answerYes{} 
    \item[] Justification: We provide many experimental details in both the main body and the appendix.

\item {\bf Experiment statistical significance}
    \item[] Question: Does the paper report error bars suitably and correctly defined or other appropriate information about the statistical significance of the experiments?
    \item[] Answer: \answerNo{} 
    \item[] Justification: Actually, the answer is partially yes and partially no. For the mutual information measuring experiment, we provide standard variance (in \Cref{fig:mech-mi}), since this method is known to be sensitive. For the other experiments, we do not repeat them under different seeds, as it would be too computationally expensive and cost too much time.

\item {\bf Experiments compute resources}
    \item[] Question: For each experiment, does the paper provide sufficient information on the computer resources (type of compute workers, memory, time of execution) needed to reproduce the experiments?
    \item[] Answer: \answerYes{} 
    \item[] Justification: See \Cref{sec:computation-resource}.
    
\item {\bf Code of ethics}
    \item[] Question: Does the research conducted in the paper conform, in every respect, with the NeurIPS Code of Ethics \url{https://neurips.cc/public/EthicsGuidelines}?
    \item[] Answer: \answerYes{} 
    \item[] Justification: We conform the NeurIPS Code of Ethics in this study.

\item {\bf Broader impacts}
    \item[] Question: Does the paper discuss both potential positive societal impacts and negative societal impacts of the work performed?
    \item[] Answer: \answerNA{} 
    \item[] Justification: This is a general AI study, and its societal impact does not go beyond that of a standard AI paper.
    
\item {\bf Safeguards}
    \item[] Question: Does the paper describe safeguards that have been put in place for responsible release of data or models that have a high risk for misuse (e.g., pre-trained language models, image generators, or scraped datasets)?
    \item[] Answer: \answerNA{} 
    \item[] Justification: This work is not likely to be exploited for misusing.

\item {\bf Licenses for existing assets}
    \item[] Question: Are the creators or original owners of assets (e.g., code, data, models), used in the paper, properly credited and are the license and terms of use explicitly mentioned and properly respected?
    \item[] Answer: \answerYes{} 
    \item[] Justification: See \Cref{sec:app-license}.

\item {\bf New assets}
    \item[] Question: Are new assets introduced in the paper well documented and is the documentation provided alongside the assets?
    \item[] Answer: \answerNA{} 
    \item[] Justification: We do not release new assets.

\item {\bf Crowdsourcing and research with human subjects}
    \item[] Question: For crowdsourcing experiments and research with human subjects, does the paper include the full text of instructions given to participants and screenshots, if applicable, as well as details about compensation (if any)? 
    \item[] Answer: \answerNA{} 
    \item[] Justification: This study does not involve crowdsourcing nor research with human subjects.

\item {\bf Institutional review board (IRB) approvals or equivalent for research with human subjects}
    \item[] Question: Does the paper describe potential risks incurred by study participants, whether such risks were disclosed to the subjects, and whether Institutional Review Board (IRB) approvals (or an equivalent approval/review based on the requirements of your country or institution) were obtained?
    \item[] Answer: \answerNA{} 
    \item[] Justification: The paper does not involve crowdsourcing nor research with human subjects.

\item {\bf Declaration of LLM usage}
    \item[] Question: Does the paper describe the usage of LLMs if it is an important, original, or non-standard component of the core methods in this research? Note that if the LLM is used only for writing, editing, or formatting purposes and does \emph{not} impact the core methodology, scientific rigor, or originality of the research, declaration is not required.
    \item[] Answer: \answerNA{} 
    \item[] Justification: We use LLMs only for writing assistance.

\end{enumerate}

\end{document}